\documentclass[acmsmall,screen]{acmart}
\usepackage{multirow}
\usepackage{geometry}
\usepackage{pdflscape}
\usepackage{enumitem}
\usepackage{soul}

\AtBeginDocument{%
  }

\begin{document}

\title{Deepfake Detection: A Comparative Analysis}

\author{Sohail Ahmed Khan}
%\authornote{Corresponding author}
\email{sohail.khan@uib.no}
\orcid{0000-0001-5351-2278}
\affiliation{%
  \institution{SFI-MediaFutures, University of Bergen}
  % \streetaddress{P.O. Box 1212}
  \country{Norway}
}

\author{Duc-Tien Dang-Nguyen}
\orcid{0000-0002-2761-2213}
\affiliation{%
  \institution{University of Bergen}
  % \streetaddress{1 Th{\o}rv{\"a}ld Circle}
  % \city{Hekla}
  \country{Norway}
  }
\email{ductien.dangnguyen@uib.no}

\renewcommand{\shortauthors}{Khan and Dang-Nguyen}

\begin{abstract}
  This paper present a comprehensive comparative analysis of supervised and self-supervised models for deepfake detection. We evaluate eight supervised deep learning architectures and two transformer-based models pre-trained using self-supervised strategies (DINO, CLIP) on four benchmarks (FakeAVCeleb, CelebDF-V2, DFDC, and FaceForensics++). Our analysis includes intra-dataset and inter-dataset evaluations, examining the best performing models, generalisation capabilities, and impact of augmentations. We also investigate the trade-off between model size and performance. Our main goal is to provide insights into the effectiveness of different deep learning architectures (transformers, CNNs), training strategies (supervised, self-supervised), and deepfake detection benchmarks. These insights can help guide the development of more accurate and reliable deepfake detection systems, which are crucial in mitigating the harmful impact of deepfakes on individuals and society.
\end{abstract}






\maketitle

\section{Introduction}

Deepfakes, or deepfake media, are digital media that have been generated or modified using deep learning algorithms. They have gained notoriety in recent years due to their potential to manipulate and deceive using artificial intelligence (AI) techniques. While deepfakes can be used for harmless or even humorous purposes, they can also pose a serious threat when used for malicious purposes such as creating convincing fake media to manipulate public opinion, influence elections, or incite violence.

The research community has been working on proposing AI-based automated systems to detect deepfakes. However, one of the major challenges in detecting deepfakes is that the deepfake generation systems are constantly evolving and improving. With the availability of cheap compute resources, and open-source software, it is becoming easier (even for people with limited technical knowledge and expertise) to create realistic deepfakes that are harder to distinguish from the real content. In addition to that, the deepfake detection, and generation is like a cat-and-mouse game~\cite{deepfakecatandmouse}, where the researchers propose detection tools by exploiting certain shortcomings of the generation systems. Soon after the release of the detection systems, the generation techniques are reinforced and made undetectable for the previously proposed detection systems by overcoming the exploited vulnerabilities. For example, in~\cite{Li2018InIO} researchers proposed a deepfake detector which exploited eye blinking as a cue to detect deepfake media (they found that deepfake faces don't blink eyes). Soon after they released their study, the newer deepfake generation algorithms generated videos which blinked eyes, and thus making the detection system useless. 

Another prominent problem of the available deepfake detection algorithms is the lack of generalisation capability. This means that the detection systems work excellently on detection deepfakes coming from the same data distribution as the training data used to train these systems. However, when exposed to deepfakes generated using different generation systems than the one which generated training samples, the detectors fail to achieve similar performance. Numerous studies have been proposed in the past which propose to employ different novel strategies to develop detection systems, however, all of the proposed studies have one thing in common, and that is, poor generalisation capability of the models on unseen data.

To address these issues, this study aims to provide insights into the challenge of detecting deepfake media by comparing multiple deep learning models and deepfake detection benchmarks. Specifically, we evaluate several different well-known image and video recognition architectures for their effectiveness in detecting deepfakes. Our primary objective is to identify which of these models perform well on unseen data as compared to other participating models.

\begin{figure*}[t]
  \centering
  \includegraphics[width=\linewidth]{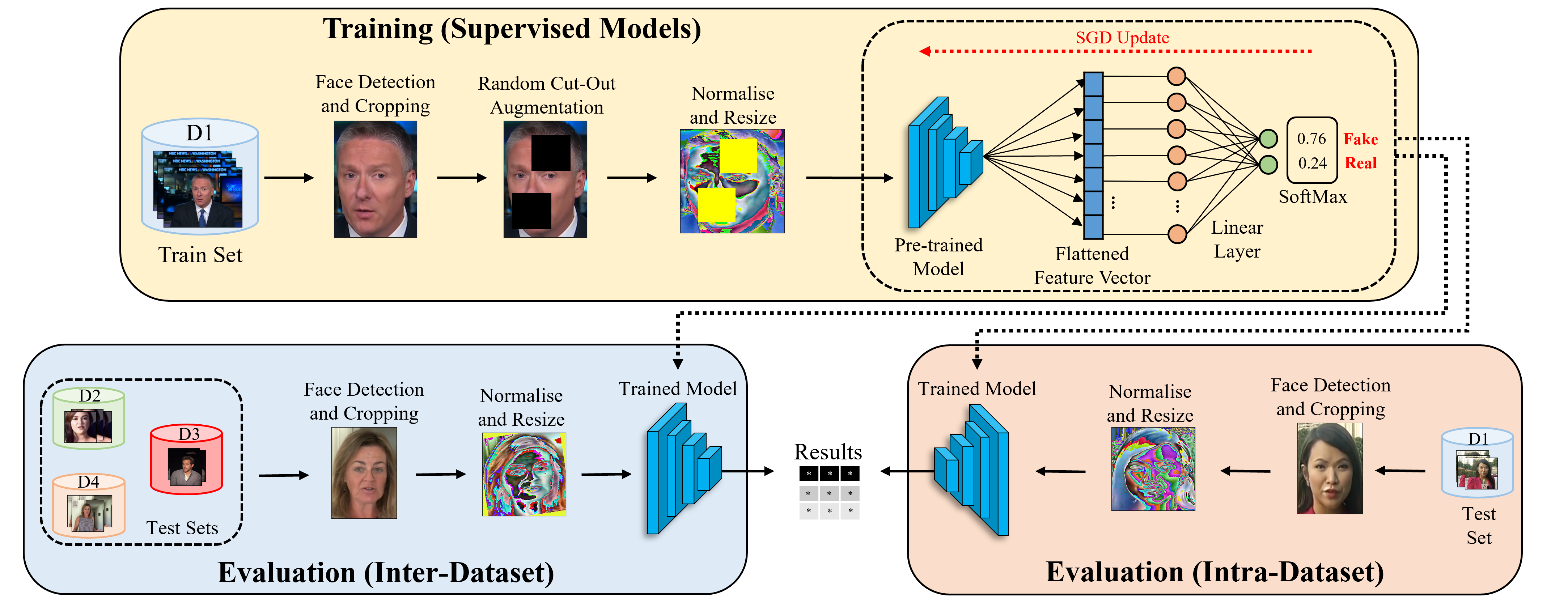}
\caption{The proposed framework. \textit{The process involves several steps, starting with the extraction and cropping of face frames from videos, followed by augmentation, normalisation, and resizing. The pre-trained models are then used as feature extractors, with a new classification head (linear layer) added on top for supervised models. During training, the weights of both the feature extractor and the classification head are updated for supervised models, while only the newly added classification head is updated for self-supervised models. The models are evaluated through both intra-dataset and inter-dataset evaluations to test their performance and generalisation capabilities. For image models, the input data is a single cropped face image, while for video models, it is a tensor containing eight consecutive cropped face images from a given video.}}
\label{fig:train_evaluate}
\end{figure*}

To achieve this, we train all participating models on four deepfake detection datasets, including a newly released dataset, and evaluate them in both intra-dataset~\footnote{models trained and evaluated on the same dataset} and inter-dataset~\footnote{models trained on one dataset and evaluated on another dataset} configurations (see Figure~\ref{fig:train_evaluate}). Additionally, we evaluate the difficulty level of each benchmark and investigate whether a more challenging benchmark leads to better generalisation performance on unseen data. To this end, we train participating models on all four datasets twice: first, without any image augmentations, and then with various image augmentations to improve their performance.

We also analyse self-supervised Vision Transformer (ViT) architectures pre-trained using two well-known strategies: \textbf{DINO}~\cite{Caron2021EmergingPI} and \textbf{CLIP}~\cite{Radford2021LearningTV}. To study these models, we use self-supervised ViT-Base models as feature extractors and train a classification head on top of them. It is important to note that we only train the classification head and freeze the weights of the feature extractors to avoid backpropagating gradients through them.

Overall, our study aims to answer several questions, such as which model has the highest generalisation capability on unseen data, which dataset is most challenging for the models to learn, which dataset enables the models to achieve the best generalisation capability on unseen data, and which of the participating models and architectures are most successful for detecting deepfakes.

This next parts of this paper is organised as follows. In Section~\ref{litreview} we present a brief literature review on the topic of deepfake detection. Section~\ref{methodology} presents the proposed framework. In Section~\ref{results} we present the results and discussion of our findings, and finally Section~\ref{conclusion} concludes this study by summarising our analysis, and presents future research direction.
%%%%%%%%%%%%%%%%%%%%%%%%%%%%%%%%%%%%%%%%%%
\section{Literature Review}
\label{litreview}
Since recently quite a large number of research studies focused on deepfake media detection have been proposed. Most of the proposed studies employ CNN models to detect deepfake media. The proposed studies also employ different strategies e.g., novel augmentation techniques, ensemble models, behavioral features, multimodal features, temporal features along with spatial information, recurrent networks, transformer models etc to detect deepfake images/videos while trying to increase the models' generalisation capabilities. Below we present some well-known, as well as some of the recently proposed deepfake detection studies.

In one of the earliest studies on deepfake media detection, Afchar~\emph{et al.} proposed two different CNN models namely (1) Meso-4, and (2) MesoInception-4~\cite{DAfchar}. Both of the proposed CNN networks were comprised of a very small number of layers which focused on mesoscopic image details. Authors tested their proposed models on one of the available deepfake detection benchmark. In addition to that, authors also collected a custom dataset and tested the models on it as well achieving excellent results on both participating datasets.

In~\cite{ESabir} Sabir~\emph{et al.} proposed to detect deepfake media using a novel recurrent convolutional network. Authors used DenseNet CNN, and combined it with a gated recurrent neural network (RNN) to learn temporal features along with spatial features. The motivation was to detect inconsistencies within neighbouring frames of a video. Authors evaluated their model on the widely known FaceForensics++~\cite{Rssler2019FaceForensicsLT} deepfake detection benchmark showing promising results.

% Following a somewhat similar direction, Guera and Delp  proposed a system comprised of a CNN along with a long short term memory (LSTM) network to detect deepfake videos~\cite{DGuera}. They proposed to exploit inter-frame discrepancies inherent to most deepfake videos. The CNN was used to extract spatial frame-level features. These features were then passed to the LSTM network to learn the temporal features. The system was evaluated on a closed deepfake detection dataset collected by the authors themselves showing eXceptional performance.

Rossler~\emph{et al.} in \cite{Rssler2019FaceForensicsLT} proposed a deepfake detection benchmark, called FaceForensics++. Along with the benchmark, authors proposed a simple CNN based deepfake detection technique using XceptionNet~\cite{Chollet2017XceptionDL}. Authors trained and evaluated the simple XceptionNet on their FaceForensics++ deepfake detection benchmark. They reported excellent performance scores on high-quality version of the four subsets of the FaceForensics++ dataset~\cite{Rssler2019FaceForensicsLT}, however, lost performance when evaluated on low-quality videos. 
 
% Qi~\emph{et al.} in~\cite{Qi2020DeepRhythmED} proposed a deepfake detection technique called, DeepRhythm. The proposed technique detected deepfakes by analysing the heartbeat rhythms of the person within a given video. DeepRhythm employed dual spatio-temporal attention in order to better cope with dynamically changing facial features over the sequence of frames. Authors evaluated the proposed technique on FaceForensics++~\cite{Rssler2019FaceForensicsLT}, and DFDC~\cite{Dolhansky2020TheDD} datasets and reported excellent performance results on both intra-dataset evaluation as well as inter-dataset evaluation as compared to other prominent deepfake detection techniques.

% In~\cite{XXuan} Xuan~\emph{et al.} proposed a deepfake media detection technique which employed image augmentations for example, gaussian blur, and gaussian noise in order to preprocess images. The augmentations helped in remove the artifacts embedded inside the GAN generated images. The resulting preprocessed images were then used to train a convolutional neural network (CNN). Authors discovered that by processing the images using the mentioned image augmentations helped their model in learning more discriminating features in both real and fake images. The proposed technique was shown to be effective during experimentation by the authors.

In~\cite{HNguyen} Nguyen~\emph{et al.} proposed to employ capsule networks for deepfake detection. The proposed technique was the first of its kind which proposed to employ capsule networks in contrast to most of the other techniques which proposed to employ CNN models at that time. The capsule networks based detection technique was evaluated on four different deepfake detection datasets comprising of a wide variety of fake videos and images. The authors reported excellent evaluation statistics of their proposed technique in comparison to other deepfake detection techniques. 

Ciftci~\emph{et al.} in~\cite{UACiftci}, developed a novel CNN and SVM based deepfake media detection model which was trained on biological signals (i.e., photoplethysmography or PPG signals). The CNN and SVM models make their individual predictions which were then fused together in order to get a final classification score. This deepfake detection model achieved promising results when tested on a number of different deepfake detection benchmarks including, CelebDF\cite{Li2020CelebDFAL}, Face Forensics, and Face Forensics++ \cite{Rssler2019FaceForensicsLT} datasets.

Zhu~\emph{et al.} in~\cite{Zhu2021FaceFD} proposed a deepfake detection system which employed 3D face decomposition features in order to detect deepfakes. Authors showed that by merging the 3D identity texture and direct light features significantly improved the detection performance while also making the model to generalise well on unseen data when evaluated in cross-dataset setting. In this study, authors also employed the XceptionNet CNN architecture for feature extraction. Both a face cropped image, and its associated 3D detail were used to train their deepfake detection model. They also carried out an in-depth analysis of several different feature fusion strategies. The proposed model was trained on the FaceForensics++~\cite{Rssler2019FaceForensicsLT} benchmark, and evaluated on (1) FaceForensics++, (2) Google Deepfake Detection Dataset, and (3) DFDC~\cite{Dolhansky2020TheDD} dataset. Promising evaluation statistics were reported for all of the three participating datasets, while depicting the generalisation capability of the model when compared to the previously proposed deepfake detection systems. 

In \cite{Khan2021VideoTF}, Khan~\emph{et al.,} proposed to employ transformer architecture for the task of deepfake media detection. Authors proposed a novel video based model for deepfake detection which was trained on 3d face features as well as standard cropped face images. Authors also showed that their proposed model was capable of incrementally learning from new data without catastrophically forgetting what it was trained on earlier. Authors evaluated their models on different widely used deepfake detection benchmarks including FaceForensics++, DFDC, DFD and showed that their proposed models achieved excellent results on all of the participating datasets.

\cite{Wang2021M2TRMM} introduce a Multi-modal Multi-scale TRansformer (M2TR) model, which processes patches of multiple sizes to identify local abnormalities in a given image at multiple different spatial levels. M2TR also utilises the frequency domain information along with RGB information using a sophisticated cross-modality information fusion block to detect forgery related artifacts in a better way. Through extensive experiments authors establish the effectiveness of M2TR, and show their model outperforms SOTA Deepfake detection models by acceptable margins.

\cite{Coccomini2021CombiningEA} propose a video deepfake detection model by employ a hybrid transformer architecture. Authors used an EfficientNet-B0 as feature extractor. The extracted features were then used to train two different types of Vision Transformer models in their study, e.g., (1) Efficient ViT, and (2) Convolutional Cross ViT. Through experimentation, authors established that the model comprising of EfficientNet-B0 feature extractor and Convolutional Cross ViT achieved the best performance scores as compared to other models that they tested.

% In \cite{Khormali2022DFDTAE}, an End-to-End DeepFake Detection Framework Using Vision Transformer (DFDT) was porposed. DFDT makes use of the unique properties of transformer models to learn traces of perturbations from both local spatial image features as well as global relationship between pixels at different forgery scales. The framework comprises of four main parts i.e., (1) patch extraction and embedding, (2) multi-stream transformer block, (3) attention-based patch selection, and (4) a multi-scale classifier. In addition to that, DFDT’s transformer block uses a re-attention mechanism instead of conventional multi-headed self-attention layer (MHSA). The DFDT was evaluated through a comprehensive set of tests on several deepfake detection benchmarks, and showed to achieve excellent performance scores on all the benchmarks.

In \cite{Zhao2023ISTVTIS}, an Interpretable Spatial-Temporal Video Transformer (ISTVT) for deepfake detection was proposed. The proposed model incorporates a novel decomposed spatio-temporal self-attention as well as a self-subtract mechanism to learn forgery related spatial artifacts and temporal inconsistencies. ISTVT can be also visualise the discriminative regions for both spatial and temporal dimensions by using the relevance propagation algorithm~\cite{Zhao2023ISTVTIS}. Extensive experiments on large-scale datasets were conducted, showing a strong performance of ISTVT both in intra-dataset and inter-dataset deepfake detection establishing the effectiveness and robustness of proposed model.

Through this literature review it becomes apparent that the research community actively employs deep learning based models along with other techniques to try develop robust and efficient deepfake detectors. However, while carefully reading the research studies it  also becomes noticeable that the models perform poorly on unseen data. Also, there is a lack of comparative studies which aim to identify which specific family of deep learning architectures is better than the others in detecting deepfakes. Also, it is a bit difficult to make sense of the capability of datasets in providing the generalisation capability to the models which can help them classify unseen data in a better way. To address this, in this study we propose to employ some of the most frequently used architectures (EfficentNets, Xception, Vision Transformers) in the literature of deepfake detection. We also employ widely known datasets for experimentation, and try to find out the datasets offering best generalisation capabilities to the models. We also analyse somewhat understudied approaches for deepfake detection i.e., we train and evaluate the performance of self-supervised models and compare their performance with supervised models.
%%%%%%%%%%%%%%%%%%%%%%%%%%%%%%%%%%%%%%%%%%
\section{The proposed framework}
\label{methodology}
The workflow followed in this study for training and evaluating the models is illustrated in Figure~\ref{fig:train_evaluate}. On top we show the training pipeline where we start by extracting and cropping faces from videos. The cropped face frames are then augmented, normalised, and resized before being fed to the model for training. We load pre-trained models as feature extractors, i.e., we remove the last layer from the loaded models and add a new classification head (linear layer) on top. For supervised models, during training we update weights of both feature extractor as well as the classification head. In case of self-supervised models we only update weights of the newly added classification head, and keep the weights of feature extractor frozen.

For intra-dataset evaluation we evaluate models on the same dataset (test set) it was trained on, e.g., model trained on dataset D1 is evaluated on the test set of D1. The main goal of intra-dataset evaluation is to find out which model performs better than other participating models on each of the dataset. Besides this, intra-dataset evaluation provides an insight about the dataset which is more challenging for the models to learn, and which dataset is the least challenging to learn.

During inter-dataset evaluation, we evaluate models trained on one dataset on each of the other three datasets, e.g., model trained on dataset D1 is evaluated on D2, D3, and D4 datasets. The goal of inter-dataset evaluation is to study the generalisation capabilities of models as well as to find out how good the training dataset is at providing that generalisation capability.

The input data for training and evaluating models image models is a single face cropped image ($[3, 224, 224]$), whereas, input data for training and evaluating video models is a tensor containing 8 consecutive face cropped images ($[8, 3, 224, 224]$) from any given video.

\subsection{Datasets}
In this study we train and evaluate several different deep learning models on four deepfake detection datasets/benchmarks: FakeAVCeleb~\cite{Khalid2021FakeAVCelebAN},
CelebDF-V2~\cite{Li2020CelebDFAL},
DFDC~\cite{Dolhansky2020TheDD}, and
FaceForensics++~\cite{Rssler2019FaceForensicsLT}. All of the four datasets comprise of real and fake videos, where fake videos are generated using different deepfake generation methods. %Three (FaceForensics++, CelebDF-V2, DFDC) of them chmarks are well-known in this field and used in most of the prominent deepfake detection research studies. FakeAVCeleb dataset which we also employ to train and evaluate our models in this study is fairly new and has not been widely studied. 
In upcoming sections, we present a brief description of these datasets.

\textbf{FaceForensics++}~\cite{Rssler2019FaceForensicsLT} is one of the most widely studied deepfake detection benchmarks. FaceForensics++ comprises of 1000 real video sequences (mostly from YouTube) of mostly frontal faces and without any occlusions. These real videos were then manipulated using four different face manipulation methods: (1) FaceSwap~\cite{faceswapKowalski}, (2) Deepfakes~\cite{deepfakesThies}, (3) Face2Face~\cite{Thies2016Face2FaceRF}, and (4) NeuralTextures~\cite{Thies2019DeferredNR} to have four subsets. 
Each subset comprises of 1000 videos each. In total, the dataset contains 5000 videos, i.e., 1000 real and 4000 fake videos. FaceForensics++ offers 3 different qualities of data, (1) Raw, (2) High-Quality and (3) Low-Quality. In our study, we experimented the high-quality videos.

\begin{figure*}[t!]
  \centering
  \includegraphics[width=0.8\linewidth]{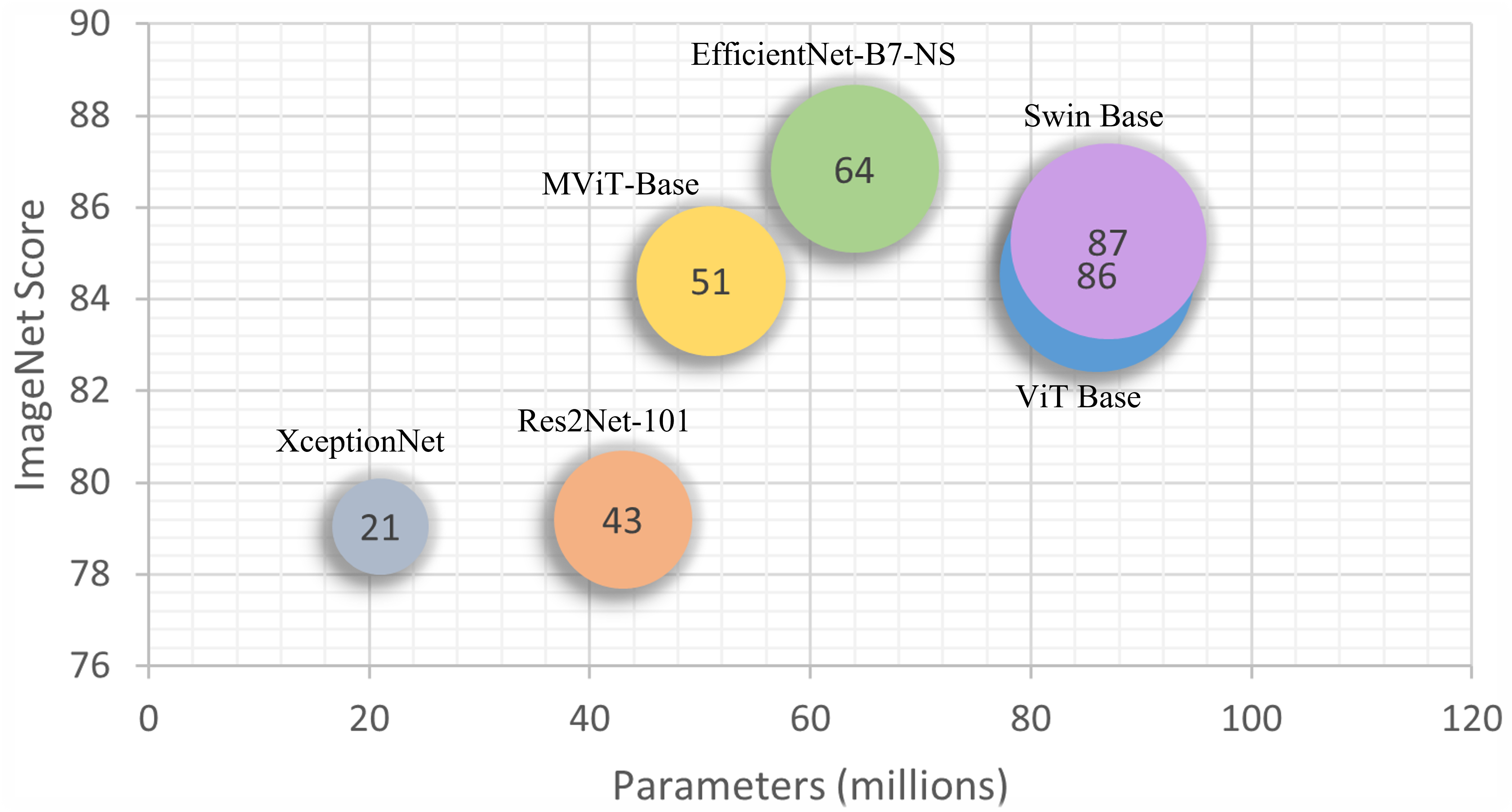}
\caption{Model size and its performance (Top-1 accuracy) on ImageNet~\cite{Deng2009ImageNetAL}.}
\label{fig:imagenetscores}
\end{figure*}

FaceSwap and Deepfakes subset contains videos generated using what is called, face swapping. As the name suggests, face of the target person is replaced with the face of source person and results in transferring the identity of the source person onto the target. Face2Face and NeuralTextures subsets are generated by a different process called, face re-enactment. In contrast to face swapping, face re-enactment swaps the faces of source and target, however, keeps the original identity of the target face (see Figure~\ref{fig:imagenetscores}).  

\textbf{Deepfake Detection Challenge (DFDC)}~\cite{Dolhansky2020TheDD} comprises of around 128k videos, out of which, around 104k are fake. Similar to the FaceForensics++ dataset, DFDC dataset also comprises of videos generated using more than one face manipulation algorithms. Five different methods were employed to generate fake videos, namely, (1) Deepfake Autoencoder\cite{Dolhansky2020TheDD}, (2) MM/NN\cite{Huang2012FacialAT}, (3) NTH\cite{Zakharov2019FewShotAL}, (4) FSGAN\cite{Nirkin2019FSGANSA}, and (5) StyleGAN\cite{Karras2018ASG}. In addition to these, a random selection of videos also underwent a simple sharpening post-processing operation which increases the videos' perceptual quality. Unlike FaceForensics++ dataset, the DFDC dataset also contains videos having having undergone audio-swapping. However, in this study we do not use audio features to train and evaluate our models. 

Since DFDC dataset is huge and we have limited resources, we only use a subset of the full dataset to train and evaluate our models. For training we use roughly around 19.5K (around 16.5K fake, and 3.1K real) videos from which we get 100k face cropped images (50k real, and 50k fake). We use 20k images as validation set. For testing the models we use 4000 image frames randomly selected from 3.5K (3.2K fake, and 0.3K real) videos.

\textbf{CelebDF-V2}~\cite{Li2020CelebDFAL} contains 5639 fake, and 590 real videos. The real videos are collected from YouTube, and contain interview videos of 59 celebrities having diverse ethnic backgrounds, genders, age groups. CelebDF-V2 dataset comprises of fake videos generated using Encoder-Decoder models. Post-processing operations are also employed to circumvent color mismatch, temporal flickering, and inaccurate face masks. 

\textbf{FakeAVCeleb}~\cite{Khalid2021FakeAVCelebAN} is the most recently proposed deepfake detection dataset. FakeAVCeleb dataset contains 19500 fake, and 500 real videos. This dataset also contains audio modality, and manipulates audio as well as video content to generated deepfake videos. For video manipulation, FaceSwap\cite{Korshunova2016FastFU}, and FSGAN\cite{Nirkin2019FSGANSA} alogrithms are used. For audio manipulation, a real-time voice cloning tool called SV2TTS\cite{Jia2018TransferLF}, and Wav2Lip\cite{KR2020ALS} are used. The dataset is divided into 4 subsets, i.e., (1) FakeVideo/FakeAudio, (2) RealVideo/RealAudio, (3) FakeVideo/RealAudio, and (4) RealVideo/FakeAudio. 

In this study, we only employ 2 of the mentioned subsets to train our models, i.e., (1) FakeVideo/FakeAudio, and (2) RealVideo/RealAudio.

\begin{table}[htbp]
  \centering
  \caption{The amount of \texttt{real/fake} images used to train, validate, and test our image models.}
  \resizebox{.7\linewidth}{!}{%
    \begin{tabular}{l|r|r|r|r|r|r}
    \toprule
    \rowcolor[rgb]{ 0,  0,  0} \multicolumn{7}{c}{\textcolor[rgb]{ 1,  1,  1}{\textbf{Train/Test Data}}} \\
    \midrule
    \multicolumn{1}{c|}{\multirow{2}[4]{*}{\textbf{Dataset}}} & \multicolumn{2}{c|}{\textbf{Train}} & \multicolumn{2}{c|}{\textbf{Validation}} & \multicolumn{2}{c}{\textbf{Test}} \\
\cmidrule{2-7}          & \multicolumn{1}{c|}{\textbf{Real}} & \multicolumn{1}{c|}{\textbf{Fake}} & \multicolumn{1}{c|}{\textbf{Real}} & \multicolumn{1}{c|}{\textbf{Fake}} & \multicolumn{1}{c|}{\textbf{Real}} & \multicolumn{1}{c}{\textbf{Fake}} \\
    \midrule
    FakeAVCeleb~\cite{Khalid2021FakeAVCelebAN} & 47,099 & 45,912 & 9,301 & 9,301 & 2,000 & 2,000 \\
    \midrule
    CelebDF-V2~\cite{Li2020CelebDFAL} & 50,000 & 50,000 & 10,000 & 10,000 & 1,000 & 1,000 \\
    \midrule
    DFDC~\cite{Dolhansky2020TheDD}  & 50000 & 50,000 & 10,000 & 10,000 & 2,000 & 2,000 \\
    \midrule
    FaceForensics++~\cite{Rssler2019FaceForensicsLT} & 50000 & 50,000 & 10,000 & 10,000 & 2,000 & 2,000 \\
    \bottomrule
    \bottomrule
    \end{tabular}%
    }
  \label{tab:dataset-statistics}
\end{table}

\subsection{Dataset Preparation}

Data preparation process was quite lengthy process as (1) the datasets are quite big, and (2) some of the chosen datasets do not come with helpful dataset preparation instructions, e.g., FakeAVCeleb does not comes with train/validation/test splits. We thus have to manually devise strategy in order to properly divide dataset into train/validation/test sets, while making sure that we do not encounter same identity in more than one splits.

Besides this, all of the datasets are unbalanced, i.e., contain more "fake" videos as compared to the "real" videos. We also made sure that we extract faces from videos in a way that the resulting face cropped image datasets are balanced, and contain at least a frame from all of the videos we chose to train/evaluate our models. 

\subsection{Preprocessing and Augmentations}

We employ two different strategies to train our models in this study. First, we train models without using any image augmentations, and second, we train the models using different randomly selected image augmentations, including, horizontal flips, affine transformations, and random cut-out augmentations. All the face cropped images are then normalised according to the same strategy used in order to pre-train models on ImageNet dataset. We use ImgAug~\cite{imgaug} library for the augmentations.

\subsection{Models}
%In this paper, we present a comparative analysis of 10 different deep learning models, including CNNs, and Transformers on the task of deepfake detection on four different benchmarks. 
We choose to experiment with six image recognition models trained using supervised strategy, three of them are CNNs and the rest three are transformer based models. We also evaluate two variants of transformer models trained using self-supervised strategies including (1) DINO~\cite{Caron2021EmergingPI}, and (2) CLIP~\cite{Radford2021LearningTV}. Besides the image classification models, we also train and evaluate two different video classification models, (1) ResNet-3D~\cite{Hara2017LearningSF}, which is a CNN model for video classification, and (2) TimeSformer~\cite{Bertasius2021IsSA}, which is a transformer model for video classification. 

We select models based on their performance on the ImageNet dataset~\cite{Deng2009ImageNetAL}, the number of parameters of the model, and for some models such as the Xception~\cite{Chollet2017XceptionDL}, we consider their previously reported performance on the task of deepfake detection. %We will further elaborate on the selection criteria in the upcoming paragraphs.
    
\begin{figure*}[t!]
  \centering
  \includegraphics[width=\linewidth]{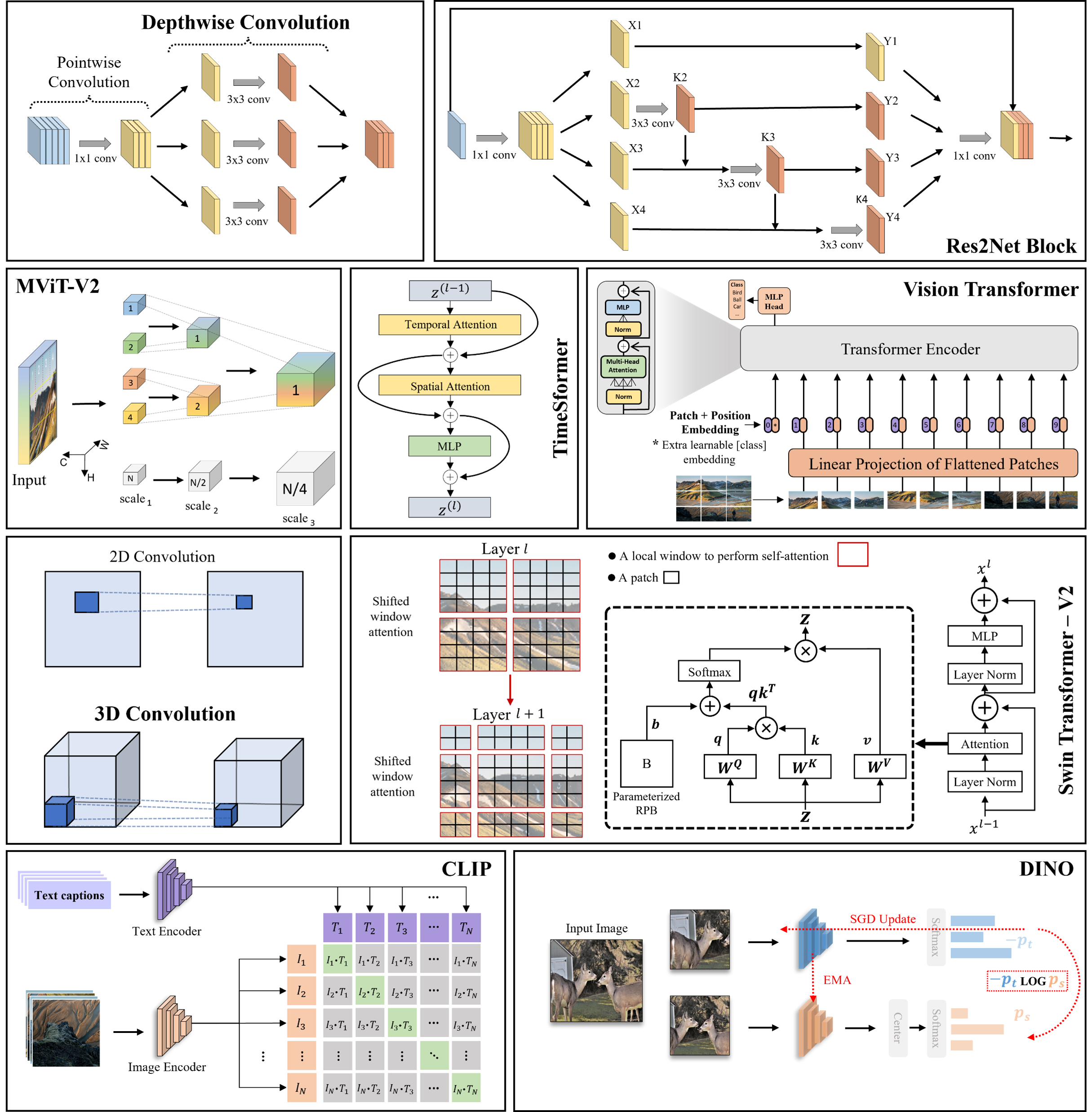}
\caption{Visual representation of the models used for analysis in this study. Due to space limitations, only basic, key concepts for each model are illustrated instead of the whole model. For optimal understanding of the essential components of each model, we recommend viewing this figure in color and at a higher magnification.}
\label{fig:allmodels}
\end{figure*}

\subsubsection{Image Models}
Deepfake detection task is typically considered as an image classification task, where a deep learning model is trained and evaluated on images separately (image-by-image), which is in contrast to the video based deepfake detection where the models are trained and evaluated on consecutive video frames to better detect the temporal inconsistencies present between different frames of the video along with spatial cues. Whereas the image based deepfake detection models only focus on learning to detect the spatial inconsistencies present in the images.

%In this section we describe some of the image based models we used to detect deepfakes, which are presented in detail in the paragraphs below.

\begin{itemize}[leftmargin=*, wide=0pt]
    \item \textbf{Xception}~\cite{Chollet2017XceptionDL} is a convolutional neural network (CNN) architecture built upon the Inception architecture~\cite{Szegedy2014GoingDW}, but proposes to use depth-wise separable convolutions instead of the traditional Inception modules. Xception has a smaller number of trainable parameters as compared to some of the other widely used deep CNN models, however, it still shows comparable performance to other models having more parameters on the ImageNet benchmark~\cite{Deng2009ImageNetAL}. Due to the smaller number of parameters, Xception is less prone to over-fitting on unseen data, as well as performs smaller number of computations, thus resulting in more efficient models. Depth-wise convolution is depicted in the top left corner of Figure~\ref{fig:allmodels}.
    Besides good performance on ImageNet benchmark, Xception was also shown to achieve excellent results on deepfake detection task in studies conducted in the past~\cite{Rssler2019FaceForensicsLT, Zhu2021FaceFD}. Based on the results achieved by this architecture in the past on deepfake detection we opt to analyse this architecture in this study as well.
        
    \item \textbf{Res2Net-101}~\cite{Gao2021Res2NetAN} is a convolutional neural network architecture. The main motivation behind Res2Net is to improve upon the popular ResNet architecture~\cite{He2016DeepRL} by introducing a new type of building block called the "Res2Net Block", replacing the traditional bottleneck residual blocks of ResNet. The Res2Net architecture represents multi-scale features at a granular level and increases the range of receptive fields of each of the network layers. This results in a more efficient and powerful network that can achieve better performance on a wide range of computer vision tasks, such as, image classification, segmentation and object detection~\cite{Gao2021Res2NetAN}. The proposed Res2Net block can be easily incorporated into other state-of-the-art backbone CNN models, e.g., ResNet\cite{He2016DeepRL}, DLA\cite{Yu2017DeepLA}, BigLittleNet\cite{Chen2018BigLittleNA}, and ResNeXt\cite{Xie2016AggregatedRT}. Res2Net block is illustrated in the top right corner of Figure~\ref{fig:allmodels}.
    We employ Res2Net-101 in this study to analyse whether multi-scale CNN features improve deepfake detection performance, and if so, does it also improves cross-dataset performance (generalisation capability)?
    
    \item \textbf{EfficientNet-B7}~\cite{Tan2019EfficientNetRM} is a convolutional neural network (CNN) architecture. The main idea behind the EfficientNet architecture is to increase the efficiency of convolutional neural networks by scaling the model's architecture, and parameters in a systematic manner. The authors proposed a new scaling technique that uniformly scales the depth, width, and resolution using a straightforward yet highly effective compound coefficient. In simple words, instead of arbitrarily scaling up model width, depth or resolution, the compound scaling strategy uniformly scales each dimension with a certain fixed set of scaling coefficients. Using this method the authors proposed seven different models of various scales~\cite{Tan2019EfficientNetRM}. The EfficientNet architecture achieves SoTA performance on a number of image classification benchmarks while being more computationally efficient than other architectures such as ResNet and Inception~\cite{Tan2019EfficientNetRM}. 
    As is the case with Xception, a variant of EfficientNet architecture, specifically the EfficientNet-B7 architecture was also shown to perform excellently on deepfake detection task. The winning solution of the Google sponsored Deepfake Detection Challenge (DFDC) was also based on these EfficientNet-B7 models~\cite{Dolhansky2020TheDD}. We thus choose to study this model in this paper.
    
    \item \textbf{Vision Transformer (ViT Base)}~\cite{Vaswani2017AttentionIA} is a class of neural network architectures based on the transformer architecture, which was initially designed for natural language processing tasks. In the context of computer vision, the Vision Transformer or simply ViT was the first transformer based architecture to be made available for image classification task~\cite{Dosovitskiy2021AnII}. It uses the self-attention mechanism to process visual data. The ViT uses a simple yet powerful approach, which is to divide the image into small patches and feed them into a transformer model at once. The small patches are then assigned positional embeddings in order to have an idea of the position of the image patch in the original image. A classification token is then inserted at the start of this input, which is then processed by the transformer encoder (similar to the encoder used in text related transformer models). The model learns to attend to different patches of the image at the same time to make predictions. By doing this, the network tends to better capture the context and relationships between different parts of the image, leading to comparable performance as compared to the SOTA CNN models on the ImageNet dataset after being trained on huge datasets, such as, the ImageNet-21k~\cite{Deng2009ImageNetAL} or the JFT-300M~\cite{Sun2017RevisitingUE} image datasets. ViT architecture is presented in Figure~\ref{fig:allmodels}, second row on the right side.
    In this study, we train and evaluate the base version of vision transformer (ViT-Base) model on the task of deepfake detection and compare its performance with other participating models.
    
    \item \textbf{Swin Transformer (Swin Base)}~\cite{Liu2021SwinTH} is a class of Vision Transformer models. It generates hierarchical feature maps by combining image patches in deeper layers. It is computationally efficient as compared to other vision transformer models, as it only performs self-attention within each local window, resulting in linear computation complexity depending on the size of the input image. In contrast, vanilla Vision Transformers produce feature maps of a single low resolution and have quadratic computation complexity to the size of the input image, due to global self-attention computation. 
    Swin Transformer achieves comparable performance when compared with other SoTA image classification models such as the EfficientNets\cite{Tan2019EfficientNetRM}. Besides image classification, Swin Transformers also perform well on tasks such as image segmentation, object detection. Figure~\ref{fig:allmodels}, third row on the right illustrates window generation, and attention calculation of Swin transformer.
    Because of the excellent performance swin transformer achieve on ImageNet, we use it for the task of deepfake detection, and try to study how it performs as compared to other participating models.

    \item \textbf{Multiscale Vision Transformer (MViT-V2 Base)} ~\cite{Fan2021MultiscaleVT} is another class of vision transformer model. Unlike traditional vision transformers, the MViTs have multiple stages that vary in both channel capacity and resolution. These stages create a hierarchical pyramid of features, where initial shallow layers focus on capturing low-level visual information with high spatial resolution, while deeper layers extract complex, high-dimensional features at a coarser spatial resolution. This approach allows the network to capture the context and relationships between different parts of the image in a better way, which results in improved performance on a broad range of computer vision tasks including image classification, image segmentation. A broad overview of the architecture of MViT is shown on the left side, second row in Figure~\ref{fig:allmodels}.
    Since MViTs are relatively new and achieve excellent performance on different vision tasks, we employ these in our study to analyse how well they perform on the task of deepfake detection.

    \item \textbf{DINO}\cite{Caron2021EmergingPI} is a simple self-supervised training method, which is interpreted as a form of self-\textit{\textbf{DI}}stillation with \textit{\textbf{NO}} labels. The authors adapted self-supervised methods to train ViT (vision transformer)~\cite{Dosovitskiy2021AnII} architecture, and ViTs trained using supervised strategies. The authors make the following observations in their study, i.e., (1) self-supervised ViT features incorporate explicit information useful for computer vision tasks such as semantic segmentation, which does not come along as evidently with supervised ViTs, and also not with CNNs; (2) self-supervised ViT features are also shown to achieve excellent performance when tested as k-NN classifiers, attaining 78.3\% top-1 on ImageNet with a ViT-small architecture. For more details about the strategy, please see in~\cite{Caron2021EmergingPI}. The DINO training strategy is shown in bottom right of Figure~\ref{fig:allmodels}. 
    Inspired from these findings, we also employ ViT-Base\cite{Dosovitskiy2021AnII} architecture trained using DINO~\cite{Caron2021EmergingPI}. In our study, we use the ViT-Base as feature extractor, and add a classification head on top. We only train the added classification head on participating deepfake detection datasets, while freezing the weights of the ViT-Base feature extractor, i.e., we do not train the feature extractor, but only the classification head.
    
    \item \textbf{Contrastive Language-Image Pre-Training (CLIP)}~\cite{Radford2021LearningTV} is a neural network that has been trained on a diverse set of (image, text) pairs in a self-supervised manner. It has the ability to infer the most suitable text excerpt for a given image using natural language, without explicit supervision for this task. It exhibits zero-shot capabilities similar to the ones exhibited by GPT-2/GPT-3. In CLIP’s original research paper, authors show that it achieves performance scores equivalent to the original ResNet50 CNN model\cite{He2016DeepRL} when evaluated on ImageNet\cite{Deng2009ImageNetAL} in a "zero-shot" fashion, i.e., even though CLIP does not use any of the 1.28 million labelled examples from the original dataset it achieves comparable performance as a ResNet50 model trained on ImageNet in a supervised manner. CLIP is illustrated in the bottom left corner in Figure~\ref{fig:allmodels}. For more details on CLIP, we refer readers to~\cite{Radford2021LearningTV}.
    We employ a ViT-Base model trained using CLIP as a feature extractor for our study. Similar to DINO, we add a classification head on top of ViT-Base trained using CLIP. For our analysis, we only train the classification head, and keep the CLIP ViT-Base features frozen i.e., we do not update its weights during training.
\end{itemize}

\subsubsection{Video Models}
We studied two different video classification models, (1) ResNet-3D (a CNN based video classifier), and (2) TimeSformer (a transformer based video classification model). We study the performance of both these models on intra, as well as inter dataset performance on four well-known deepfake detection benchmarks. We choose to study video based models in addition to the image based detection models in order to find out whether temporal information help in the detection task. Below we briefly describe these models.

\begin{itemize}[leftmargin=*, wide=0pt]
    \item \textbf{ResNet-3D}~\cite{Hara2017LearningSF} is based on the same principles as the original ResNet architecture~\cite{He2016DeepRL}, but they are specifically designed to work with 3D data, such as videos and volumetric medical images. These models use 3D convolutions, instead of 2D layers, for feature extraction. In addition to that, ResNet-3D models generally use a large number of layers, which allows them to learn complex and abstract features in the data. ResNet-3D models have been utilised for a variety of computer vision tasks, including video classification, action recognition, and medical image segmentation. They have been shown to achieve SoTA performance on a number of different benchmarks, however, it is also worth noting that the ResNet-3D models are computationally costly, and need a large amount of data to train. For reference, we illustrate both 2D and 3D convolutions in Figure~\ref{fig:allmodels}, on the left side of third row.
    We choose to employ ResNet-3D model for our study because, (1) it is widely studied in regards of video recognition, (2) pre-trained models are available, and (3) the available compute power which is not suitable for training bigger video recognition models using more video frames for training. We chose ResNet-3D model pre-trained on 8 frames per video to experiment in this study. In contrast, available video classification models are typically trained on 16/32 frames per video and tend to perform better than models trained using 8 frames per video. 
    
    \item \textbf{TimeSformer}~\cite{Bertasius2021IsSA} is a video recognition model based on the transformer architecture. TimeSformer utilises self-attention over space and time, instead of traditional convolutional layers, or the spatial attention as employed by ViT for image recognition. The TimeSformer model modifies the transformer architecture, generally used for image recognition, by directly learning the spatio-temporal features from a sequence of frame-level patches. This is accomplished by extending the self-attention mechanism from the image space to the 3D space-time volume. Similar to the Vision Transformer (ViT) model, the TimeSformer employs linear mapping and positional information to interpret ordering of the resulting sequence of features.
    In TimeSformer paper~\cite{Bertasius2021IsSA}, authors experimented with different self-attention techniques. Out of different techniques, the "divided attention" technique which calculates temporal and spatial attention separately within each block, was found to perform better than other self-attention calculation techniques, and thus we choose to analyse the same architecture in this study. Divided space-time attention is illustrated in Figure~\ref{fig:allmodels}, in the middle of second row.
    We opt to evaluate TimeSformer on the task of deepfake detection, and compare it with convolutional video classification network, ResNet-3D. We also chose 8 frame per video version of the TimeSformer model, same as the ResNet-3D model we described above.
\end{itemize}

\subsection{Evaluation Metrics}

In order to analyse the performance of our models in a comprehensive way, we employ multiple widely used classification metrics, e.g., (1) LogLoss, (2) AUC, and (3) Accuracy. Below we briefly describe the chosen evaluation metrics.

\subsubsection{LogLoss}, also known as logarithmic loss or cross-entropy loss, is used to measure the classification performance of machine/deep learning models. LogLoss calculates the dissimilarity between the predicted probability score with the true label (0, 1 in case of binary classification). The LogLoss score is computed as the negative logarithm of the likelihood of the true labels given a set of predicted probabilities. The range of the LogLoss function is from 0 to infinity, with 0 representing the ideal outcome and higher values representing worse outcomes.

\begin{equation}
L = -\frac{1}{N} \sum_{i=1}^{N} [y_i \log(p_i) + (1-y_i) \log(1-p_i)]
\label{eq:logloss}
\end{equation}

Where $L$ is the LogLoss, $N$ is the total number of samples in the dataset, $y_i$ is the true label of the \textit{i-th} sample, $p_i$ is the predicted probability for the \textit{i-th} sample.

It is worth noting that Logloss is a widely used evaluation metric in machine learning competitions such as Kaggle competitions, as it gives a general idea of how good the predictions of the model are. We use LogLoss as one of the evaluation metrics in this study as other previously proposed deepfake detection research studies often use it as their evaluation metric, and thus would allow us to compare our results with them.

\subsubsection{Area Under the Curve (AUC)} is also a widely known metric used to evaluate classification models. AUC basically refers to calculating the entire two-dimensional area under the Receiver Operating Curve (ROC). AUC gives hints about how well a model has made a certain prediction. Quite understandably, the higher the area falling under the ROC, i.e., AUC, the better the performance of the model at discriminating between "real" and "fake" samples in our case. Most of the recently proposed deepfake detection studies employ AUC as the evaluation metric to study the performance of their models.

Note that the ROC curve is created by varying the threshold used to make predictions from 0 to 1, so the AUC provides a summary of the model's performance across all possible thresholds.
% \begin{equation}
%  AUC = \frac{1}{2} \bigg[ \sum_{i=1}^{N^-} \sum_{j=1}^{N^+} (y_i < y_j) + \sum_{i=1}^{N^+} \sum_{j=1}^{N^-} (y_i = y_j) \bigg]
%  \label{eq:auc}
% \end{equation}

\subsubsection{Accuracy} is also a widely used metric in the classification domain. Accuracy score is basically the measure of correct predictions made by a model in relation to all the predictions made by the model. Accuracy does not indicate how well a model has made a certain classification, as was the case with LogLoss and AUC. Accuracy score can be obtained by dividing the number of correct predictions by total predictions.

\begin{equation}
Accuracy = \frac{TP + TN}{TP + FP + TN + FN}
\label{eq:acc}
\end{equation}

Where $TP$ is the number of true positives, $TN$ refers to the number of true negatives, $FP$ refers to the number of false positives, and $FN$ refers to the number of false negatives. 

It is worth noting that accuracy is the proportion of correctly classified samples out of the total number of samples. It is a common evaluation metric used in binary classification tasks, however, it can be misleading in cases where the classes (real, fake) are imbalanced, or if the cost associated with the false positives and false negatives is different. In such cases, other evaluation metrics like F1 score, precision, recall, or AUC may provide a more accurate evaluation of the classification model's performance~\cite{accuracyparadox}. In our study however, since we have balanced number of samples both for \textbf{real} and \textbf{fake} classes, we can use accuracy as one of the evaluation metric.

\subsection{Implementation Details}

We use PyTorch library to train and test our the models. To train our models we choose a batch size of 16 for image models, and 4 for video models. We choose a constant learning rate of $3\times 10^{-3}$ for both image and video models. We use CrossEntropyLoss as the loss function and SGD (Stochastic Gradient Descent) as the optimiser to train our models. We train our models for 5 epochs, and choose the model with lowest validation loss for further testing and evaluation. For evaluation we rely on Scikit-Learn library, i.e., to report on LogLoss, AUC, Accuracy scores. 

We rely heavily on Ross Wightman's PyTorch Image Models repository~\cite{rw2019timm} for model code implementations and pre-trained weights. We adapt some code from~\cite{Caron2021EmergingPI} to train linear classification head on top of self-supervised feature extractors i.e., DINO and CLIP. For image augmentations we rely on \textit{imgaug}~\cite{imgaug} library

\begin{figure*}[!htb]
  \centering
  \includegraphics[ width=0.7\linewidth]{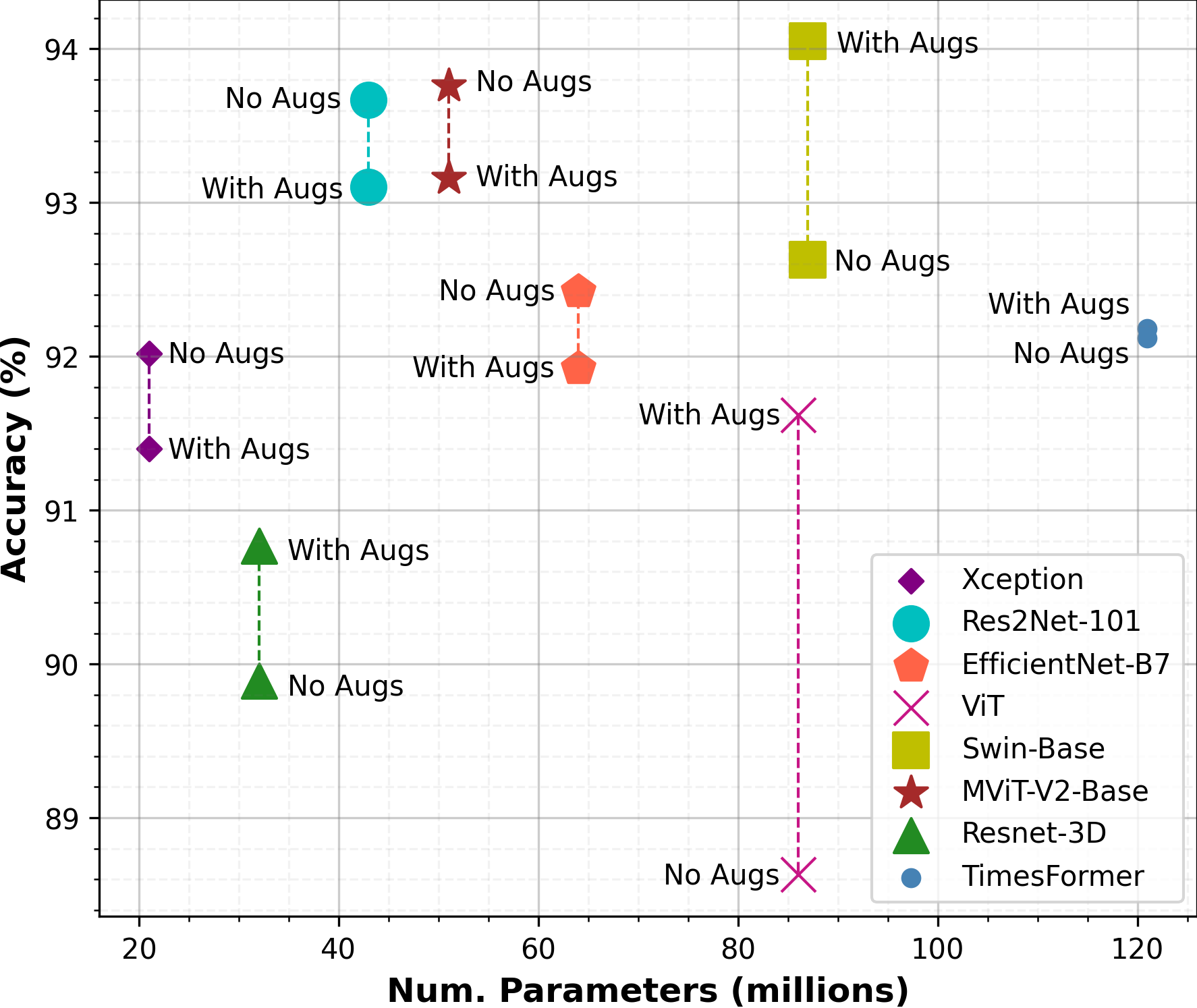}
\caption{Performance (accuracy) comparison of participating models on all datasets. The reported scores were achieved in an intra-dataset evaluation. Results in this figure are obtained by, (1) evaluating each model separately on each dataset, and (2) averaging the achieved scores i.e., add the 4 accuracy scores and divide by 4.}
\label{fig:accuracy_comparison_all}
\end{figure*}

\section{Results}
\label{results}
%Deepfake detection, i.e., whether to develop image based models, or whether to move forward with developing specialised video based models. Also, we get an idea about the model size and accuracy trade-of, and whether the transformer based architectures are better than the CNN architectures for detecting deepfakes.

We conducted extensive experiments and evaluated six image deepfake detection models, as well as two video deepfake detection models on four different benchmarks (the details were discussed in Section~\ref{methodology}). In addition to this, we also evaluate two vision transformer (ViT-Base) models pre-trained using self-supervised techniques mentioned in Section~\ref{methodology}. We evaluated all models in both intra-dataset as well as inter-dataset settings. In the following sections, we report the performance of all the participating models both in an intra-dataset (trained and evaluated on same dataset), as well as inter-dataset (trained on one dataset and evaluated on the remaining datasets excluding the training dataset) settings. %We present results in the paragraphs below for each participating dataset one by one.

In this section we refer to models as supervised models, and self-supervised models. Supervised models refer to eight models including six image models, and two video models. Self-supervised models refer to DINO, CLIP and a supervised ViT-Base (which is used as a feature extractor to compare with DINO, and CLIP based ViT-Base). Supervised models are trained end to end i.e., weights of feature extractor as well as the classification head are updated during training. In case of self-supervised models including DINO, CLIP, and a supervised ViT-Base, the weights of feature extractors are kept frozen during training and only the classification head is trained.

DINO, and CLIP are also ViT-Base models, however, the only difference is that both DINO and CLIP are pre-trained using self-supervised training strategies. The supervised ViT-Base is pre-trained using supervised training strategy. Through training a classification head on top of these three models we aim to find out whether self-supervised features provide better feature representations as compared to supervised features.

\begin{table}[h!]
  \centering
  \small
  \caption{Intra-dataset comparison of image models. The table below presents scores achieved by image models when trained and evaluated on FakeAVCeleb~\cite{Khalid2021FakeAVCelebAN} dataset. Best results are highlighted in yellow.}
  \resizebox{.75\linewidth}{!}{%
    \begin{tabular}{l|c|c|c|c|c|c}
    \toprule
    \rowcolor[rgb]{ 0,  0,  0} \multicolumn{7}{c}{\textcolor[rgb]{ 1,  1,  1}{\textbf{FakeAVCeleb}}} \\
    \midrule
    \multicolumn{1}{c|}{\multirow{2}[4]{*}{\textbf{Model}}} & \multicolumn{3}{c|}{\textbf{With Augs}} & \multicolumn{3}{c}{\textbf{No Augs}} \\
\cmidrule{2-7}          & \textbf{LogLoss} & \textbf{AUC} & \textbf{ACC} & \textbf{LogLoss} & \textbf{AUC} & \textbf{ACC} \\
    \midrule
    Xception & 0.0047 & \cellcolor[rgb]{ 1,  .753,  0}\textbf{100.00\%} & 99.93\% & 0.0040 & \cellcolor[rgb]{ 1,  .753,  0}\textbf{100.00\%} & 99.85\% \\
    \midrule
    Res2Net-101 & \cellcolor[rgb]{ 1,  .753,  0}\textbf{0.0008} & \cellcolor[rgb]{ 1,  .753,  0}\textbf{100.00\%} & 99.98\% & 0.0037 & \cellcolor[rgb]{ 1,  .753,  0}\textbf{100.00\%} & 99.93\% \\
    \midrule
    EfficientNet-B7 & 0.0132 & \cellcolor[rgb]{ 1,  .753,  0}\textbf{100.00\%} & 99.63\% & 0.0047 & \cellcolor[rgb]{ 1,  .753,  0}\textbf{100.00\%} & 99.83\% \\
    \midrule
    ViT   & 0.2073 & 99.29\% & 94.60\% & 0.3768 & 98.78\% & 92.43\% \\
    \midrule
    Swin  & 0.0033 & \cellcolor[rgb]{ 1,  .753,  0}\textbf{100.00\%} & 99.88\% & 0.0058 & \cellcolor[rgb]{ 1,  .753,  0}\textbf{100.00\%} & 99.83\% \\
    \midrule
    MViT  & \cellcolor[rgb]{ 1,  .753,  0}\textbf{0.0008} & \cellcolor[rgb]{ 1,  .753,  0}\textbf{100.00\%} & \cellcolor[rgb]{ 1,  .753,  0}\textbf{100.00\%} & \cellcolor[rgb]{ 1,  .753,  0}\textbf{0.0023} & \cellcolor[rgb]{ 1,  .753,  0}\textbf{100.00\%} & 99.95\% \\
    \midrule
    ResNet-3D & 0.0041 & \cellcolor[rgb]{ 1,  .753,  0}\textbf{100.00\%} & \cellcolor[rgb]{ 1,  .753,  0}\textbf{100.00\%} & 0.0066 & \cellcolor[rgb]{ 1,  .753,  0}\textbf{100.00\%} & \cellcolor[rgb]{ 1,  .753,  0}\textbf{100.00\%} \\
    \midrule
    TimeSformer & 0.0796 & 99.96\% & 97.50\% & 0.1238 & 99.94\% & 97.00\% \\
    \bottomrule
    \bottomrule
    \end{tabular}%
    }
  \label{tab:intra-fakeavceleb}%
\end{table}%

\subsection{FakeAVCeleb}
FakeAVCeleb~\cite{Khalid2021FakeAVCelebAN} is a newly proposed deepfake detection dataset containing four different categories of videos i.e., (1) FakeVideo/FakeAudio, (2) RealVideo/RealAudio, (3) FakeVideo/RealAudio, and (4) RealVideo/FakeAudio. Since we focus only on visual deepfakes in this study, and thus do not use the audios (real and fake) for training and evaluating our models. In fact, out of four subsets of the FakeAVCeleb dataset, we only use two for our experiments i.e., (1) FakeVideo/FakeAudio, (2) RealVideo/RealAudio.

Table~\ref{tab:intra-fakeavceleb} shows that all models perform pretty well in distinguishing between fake and real faces. We can see that all of the participating models achieved almost 99\% AUC, and very low LogLoss score when tested in an intra-dataset configuration. The numbers in~\ref{tab:intra-fakeavceleb} suggest that FakeAVCeleb dataset is relatively easy and thus the models can accurately distinguish between real and fake samples.

In table~\ref{tab:inter-fakeavceleb} we report results achieved by all the models when trained on FakeAVCeleb, and evaluated on the remaining three datasets. When we look at the numbers in Table~\ref{tab:inter-fakeavceleb}, it is apparent that almost all of the models perform poorly on all the other datasets. We can see that in terms of accuracy scores, the models are making random guesses. LogLoss, and AUC scores are also not remarkably good in inter-dataset evaluation. 

In case of self-supervised models, the results are not as good as they are for the supervised models. That is because the self-supervised models are not trained in an end to end manner as we mentioned earlier. However, on FakeAVCeleb dataset, even though only the classification head is trained, still DINO, and ViT-Base (the supervised feature extractor) achieve good performance. However, DINO performs significantly better than the other two models as shown in Table~\ref{tab:all-models-intra-self-supervised}. CLIP does not achieve good performance, and this might be because CLIP was initially pre-trained using both images and their associated text captions, however, in this study we use CLIP's image encoder only without any text captions. This might be a reason of bad performance by CLIP. We aim at investigating this issue along with the inter-dataset analysis of self-supervised models in our future research.

From the results we can infer that FakeAVCeleb dataset is not challenging enough for the models to learn, and is fairly easy to distinguish between fake and real samples for both supervised and self-supervised models. In addition to that, this dataset does not enhance the models' ability to learn distinguishing features between real and fake faces, or in other words, it lacks at integrating the generalisation capability into the models, as is apparent from Table~\ref{tab:allcomparisoninterdataset} and \ref{tab:inter-fakeavceleb}.

\begin{table}[h!]
  \centering
  \caption{Intra-dataset comparison of image models. The table below presents scores achieved by image models when trained and evaluated on CelebDF-V2~\cite{Li2020CelebDFAL} dataset.}
  \resizebox{.75\linewidth}{!}{%
    \begin{tabular}{l|c|c|c|c|c|c}
    \toprule
    \rowcolor[rgb]{ 0,  0,  0} \multicolumn{7}{c}{\textcolor[rgb]{ 1,  1,  1}{\textbf{CelebDF}}} \\
    \midrule
    \multicolumn{1}{c|}{\multirow{2}[4]{*}{\textbf{Model}}} & \multicolumn{3}{c|}{\textbf{With Augs}} & \multicolumn{3}{c}{\textbf{No Augs}} \\
\cmidrule{2-7}          & \textbf{LogLoss} & \textbf{AUC} & \textbf{ACC} & \textbf{LogLoss} & \textbf{AUC} & \textbf{ACC} \\
    \midrule
    Xception & 0.0712 & 99.73\% & 97.00\% & 0.0367 & 99.95\% & 98.55\% \\
    \midrule
    Res2Net-101 & \cellcolor[rgb]{ 1,  .753,  0}\textbf{0.0237} & \cellcolor[rgb]{ 1,  .753,  0}\textbf{100.00\%} & 98.95\% & 0.0185 & 99.99\% & 99.45\% \\
    \midrule
    EfficientNet-B7 & 0.0433 & 99.95\% & 98.40\% & 0.0340 & 99.98\% & 98.75\% \\
    \midrule
    ViT   & 0.0336 & 99.96\% & 98.60\% & 0.0350 & 99.95\% & 98.60\% \\
    \midrule
    Swin  & 0.0340 & 99.94\% & 98.80\% & 0.0202 & 99.97\% & 99.40\% \\
    \midrule
    MViT  & 0.0075 & \cellcolor[rgb]{ 1,  .753,  0}\textbf{100.00\%} & \cellcolor[rgb]{ 1,  .753,  0}\textbf{99.70\%} & \cellcolor[rgb]{ 1,  .753,  0}\textbf{0.0096} & \cellcolor[rgb]{ 1,  .753,  0}\textbf{100.00\%} & \cellcolor[rgb]{ 1,  .753,  0}\textbf{99.70\%} \\
    \midrule
    ResNet-3D & 0.0748 & 99.68\% & 97.00\% & 0.1525 & 98.68\% & 95.00\% \\
    \midrule
    TimeSformer & 0.0309 & \cellcolor[rgb]{ 1,  .753,  0}\textbf{100.00\%} & 98.00\% & 0.0220 & 99.96\% & 99.00\% \\
    \bottomrule
    \bottomrule
    \end{tabular}%
    }
  \label{tab:intra-celeb}%
\end{table}%

\subsection{CelebDF-V2}

Table~\ref{tab:intra-celeb} presents the performance of supervised models when trained and evaluated on CelebDF-V2~\cite{Li2020CelebDFAL} dataset. Same as it was the case with FakeAVCeleb dataset, almost all of the participating models achieve excellent scores i.e., more than 97\% accuracy, and more than 99\% AUC score, while having a very small LogLoss. We can thus infer that the models quite comfortably learnt to discriminate between real/fake samples of the CelebDF-V2 dataset, similar to FakeAVCeleb dataset.

In order to find out how helpful the dataset is in making models learn robust features making them better at generalisation, we also conduct extensive inter-dataset evaluation of all the participating models trained on CelebDF-V2. We report the results achieved by the models in Table~\ref{tab:inter-celeb}. However, similar to the results achieved by the models trained on FakeAVCeleb dataset and evaluated on the remaining datasets, the models trained on CelebDF-V2 and evaluated in an inter-dataset setting also seem to perform poorly. This might be a result of CelebDF-V2 not being a very challenging dataset for the models to discriminate, and they can almost classify every real/fake sample in a perfect manner. However, this also makes the models less powerful against unseen data, as can be seen by the performance scores reported in table~\ref{tab:inter-celeb}.

The fact that CelebDF-V2 is also not a challenging dataset is also supported by the results achieved by the self-supervised models as presented in Table~\ref{tab:all-models-intra-self-supervised}. It is apparent from the numbers that only training a classification head on top of frozen feature extractors, models still achieve good results. However, in this case as well, CLIP does not achieve good performance as compared to DINO and supervised ViT. 

% CelebDF-V2 dataset also found to be relatively easier for the models to discriminate between fake and real faces. All the models achieve more than 95\% AUC and relatively low LogLoss score.

\begin{table}[h!]
  \centering
  \caption{Intra-dataset comparison of image models. The table below presents scores achieved by image models when trained and evaluated on FaceForensics++~\cite{Rssler2019FaceForensicsLT} dataset.}
  \resizebox{.75\linewidth}{!}{%
    \begin{tabular}{l|c|c|c|c|c|c}
    \toprule
    \rowcolor[rgb]{ 0,  0,  0} \multicolumn{7}{c}{\textcolor[rgb]{ 1,  1,  1}{\textbf{FaceForensics++}}} \\
    \midrule
    \multicolumn{1}{c|}{\multirow{2}[4]{*}{\textbf{Model}}} & \multicolumn{3}{c|}{\textbf{With Augs}} & \multicolumn{3}{c}{\textbf{No Augs}} \\
\cmidrule{2-7}          & \textbf{LogLoss} & \textbf{AUC} & \textbf{ACC} & \textbf{LogLoss} & \textbf{AUC} & \textbf{ACC} \\
    \midrule
    Xception & 0.2342 & 96.96\% & 91.05\% & 0.2957 & 95.85\% & 89.03\% \\
    \midrule
    Res2Net-101 & 0.2165 & 97.87\% & 93.48\% & 0.3213 & 97.30\% & 91.85\% \\
    \midrule
    EfficientNet-B7 & 0.3111 & 96.92\% & 90.33\% & 0.3737 & 94.02\% & 86.95\% \\
    \midrule
    ViT   & 0.2445 & 97.27\% & 92.18\% & 0.3571 & 94.04\% & 85.15\% \\
    \midrule
    Swin  & \cellcolor[rgb]{ 1,  .753,  0}\textbf{0.1573} & \cellcolor[rgb]{ 1,  .753,  0}\textbf{98.58\%} & \cellcolor[rgb]{ 1,  .753,  0}\textbf{94.90\%} & 0.2191 & 97.60\% & 92.18\% \\
    \midrule
    MViT  & 0.1828 & 98.34\% & 94.10\% & \cellcolor[rgb]{ 1,  .753,  0}\textbf{0.1918} & \cellcolor[rgb]{ 1,  .753,  0}\textbf{97.63\%} & \cellcolor[rgb]{ 1,  .753,  0}\textbf{93.00\%} \\
    \midrule
    ResNet-3D & 0.3224 & 96.42\% & 90.36\% & 0.3085 & 96.19\% & 91.07\% \\
    \midrule
    TimeSformer & 0.2807 & 97.10\% & 90.00\% & 0.2451 & 96.76\% & 90.71\% \\
    \bottomrule
    \bottomrule
    \end{tabular}%
    }
  \label{tab:intra-ff}%
\end{table}%

\subsection{FaceForensics++}
Table~\ref{tab:intra-ff} reports the performance metrics of all the supervised models when trained and evaluated on the FaceForensics++~\cite{Rssler2019FaceForensicsLT} dataset. We can see that the results are not as good as they were in the case of previous two datasets, FakeAVCeleb and CelebDF-V2. Almost none of the models achieved more than 95\% accuracy, and the LogLoss scores are also not as great as they were for the previous two datasets. We can thus infer by looking at the metrics that it is a relatively challenging dataset for the models to distinguish between real/fake samples. Self-supervised models also are not able to achieve excellent results on FaceForensics++ dataset as is apparent from the numbers in Table~\ref{tab:all-models-intra-self-supervised}, affirming that it is indeed challenging to properly distinguish between fake and real faces coming from FaceForensics++ dataset. What we would now like to see now is, whether a more challenging dataset means better generalisation capability?

We thus evaluate all of the supervised models trained on FaceForenscis++ dataset in an inter-dataset evaluation setting and report the results in Table~\ref{tab:inter-ff}. In this case, it can be clearly seen that the models perform in a somewhat acceptable manner even on unseen data from other datasets. For example, MViT trained on FaceForenscis++ and evaluated on FakeAVCeleb dataset achieves more than 80\% accuracy, and more than 90\% AUC score. This also supports our claim that FakeAVCeleb as well as CelebDF-V2 datasets are not very challenging, and that the models can easily learn to distinguish real/fake videos coming from these datasets.

Furthermore, not only on the FakeAVCeleb dataset, we can also see somewhat better performance from all of the participating supervised models on the other two datasets, i.e., CelebDF-V2 and DFDC. However, it must be noted that how all the models suffer when tested using unseen data, i.e., lack of generalisation, which is a big problem the current, more sophisticated deepfake detection systems suffer from. The results in Table~\ref{tab:inter-ff} somehow support the statement that more challenging datasets mean better generalisation capability. But we have to further re-enforce this statement after analysing the metrics of models trained using DFDC~\cite{Dolhansky2020TheDD} dataset in the section below.

\begin{table}[t!]
  \centering
      \caption{Intra-dataset comparison of image models. The table below presents scores achieved by image models when trained and evaluated on DFDC~\cite{Dolhansky2020TheDD} dataset.}
      \resizebox{.75\linewidth}{!}{%
    \begin{tabular}{l|c|c|c|c|c|c}
    \toprule
    \rowcolor[rgb]{ 0,  0,  0} \multicolumn{7}{c}{\textcolor[rgb]{ 1,  1,  1}{\textbf{DFDC}}} \\
    \midrule
    \multicolumn{1}{c|}{\multirow{2}[3]{*}{\textbf{Model}}} & \multicolumn{3}{c|}{\textbf{With Augs}} & \multicolumn{3}{c}{\textbf{No Augs}} \\
\cmidrule{2-7}          & \textbf{LogLoss} & \textbf{AUC} & \textbf{ACC} & \textbf{LogLoss} & \textbf{AUC} & \textbf{ACC} \\
    \midrule
    Xception & 0.5613 & 88.75\% & 77.63\% & 0.5120 & 91.68\% & 80.65\% \\
    \midrule
    Res2Net-101 & 0.5570 & 90.64\% & 79.98\% & 0.5691 & 91.78\% & 83.45\% \\
    \midrule
    EfficientNet-B7 & 0.5542 & 89.97\% & 79.30\% & \cellcolor[rgb]{ 1,  .753,  0}\textbf{0.4263} & \cellcolor[rgb]{ 1,  .753,  0}\textbf{93.30\%} & \cellcolor[rgb]{ 1,  .753,  0}\textbf{84.15\%} \\
    \midrule
    ViT   & \cellcolor[rgb]{ 1,  .753,  0}\textbf{0.4696} & \cellcolor[rgb]{ 1,  .753,  0}\textbf{91.89\%} & 81.08\% & 0.5709 & 89.44\% & 78.35\% \\
    \midrule
    Swin  & 0.5602 & 90.89\% & 82.60\% & 0.6650 & 87.77\% & 79.05\% \\
    \midrule
    MViT  & 0.6079 & 88.41\% & 78.90\% & 0.5491 & 90.65\% & 82.40\% \\
    \midrule
    ResNet-3D & 0.5865 & 85.64\% & 75.75\% & 0.6739 & 84.69\% & 73.50\% \\
    \midrule
    TimeSformer & 0.4870 & 91.18\% & \cellcolor[rgb]{ 1,  .753,  0}\textbf{83.25\%} & 0.6176 & 92.30\% & 81.75\% \\
    \bottomrule
    \bottomrule
    \end{tabular}%
    }
  \label{tab:intra-dfdc}%
\end{table}%

\subsection{DFDC}

DFDC is one of the biggest and most challenging deepfake detection benchmarks. This is apparent by the results we present in Table~\ref{tab:intra-dfdc}. Only one of the supervised models (i.e., Res2Net-101) managed to achieve more than 84\% accuracy score, 93\% AUC score on the DFDC dataset. Self-supervised models also achieve relatively low scores when trained and evaluated on DFDC, as apparent from Table~\ref{tab:all-models-intra-self-supervised}. This establishes that DFDC is the most challenging dataset out of all the four datasetS in this study. 

In Table~\ref{tab:inter-dfdc} we present inter-dataset evaluation scores achieved by the supervised models trained on DFDC dataset. It is apparent from the results that the models trained using DFDC dataset still achieve acceptable performance on unseen data, as compared to the scores achieved by the models which were trained on FakeAVCeleb, and CelebDF-V2. Also, by looking at the results we can \textit{\textbf{somewhat}} affirm the statement that models trained using more challenging datasets seem to achieve better results. We say \textit{\textbf{somewhat}}, because in the scope of this study, even though DFDC is more challenging to learn for the models, a better generalisation is offered by FaceForensics++ which is relatively less challenging to learn for the models.

\subsection{Discussion}

In Figure~\ref{fig:accuracy_comparison_all} we present a comparison of all the participating models on the basis of achieved accuracies while evaluated in an intra-dataset setting (i.e., models are trained and tested on same dataset). From the figure, it is apparent that there is not a lot of performance difference between the participating models. In most cases, the models achieves around 92\% to 94\% accuracies when tested in an intra-dataset evaluation setting. The figure also shows that the image augmentations do not prove to be significantly helpful in all cases, for example, we can see that for XceptionNet, Res2Net-101, MViT-V2-Base, and EfficientNet-B7, the models trained without image augmentations achieved better scores when compared to models trained using image augmentations. The difference between the accuracies achieved by the models when trained with and without image augmentations is not big except for the ViT. The ViT trained with image augmentations achieved 91.62\% while the ViT trained without image augmentations achieved 88.63\% accuracy. However, it is apparent from the Figure~\ref{fig:accuracy_comparison_all} that all of the transformer models perform better when trained using augmentations. Besides this, video models also perform when trained using augmentations. Another reason to no disregard image augmentations is that the best performing model Swin-Base achieved top most accuracy while being trained using image augmentations.

\begin{figure*}[!h]
  \centering
  \includegraphics[width=0.8\linewidth]{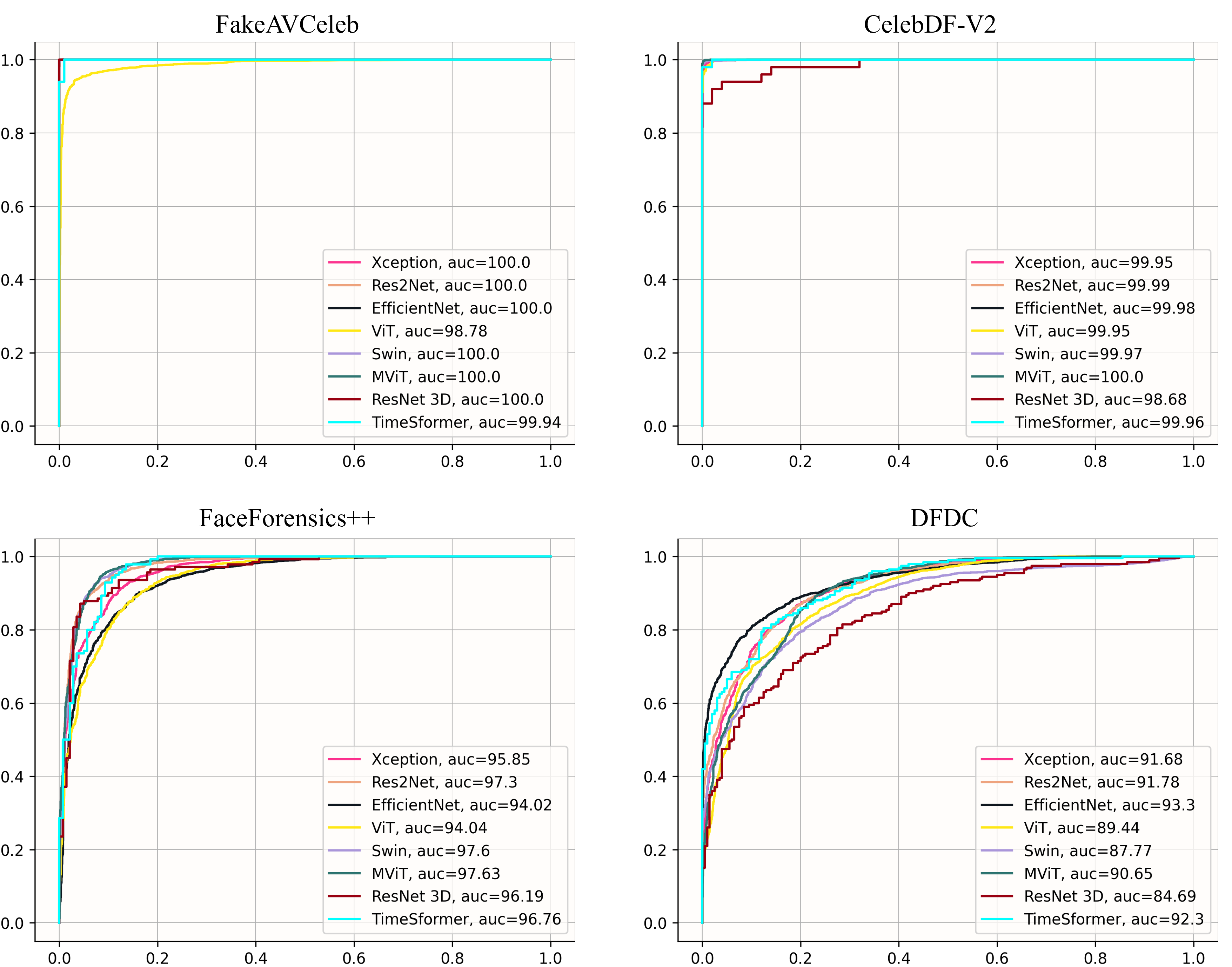}
\caption{ROC curves of each of the model when evaluated on each of the 4 different participating datasets in an intra-dataset evaluation setting.}
\label{fig:roc}
\end{figure*}

\begin{figure*}[!htb]
  \centering
  \includegraphics[width=0.85\linewidth]{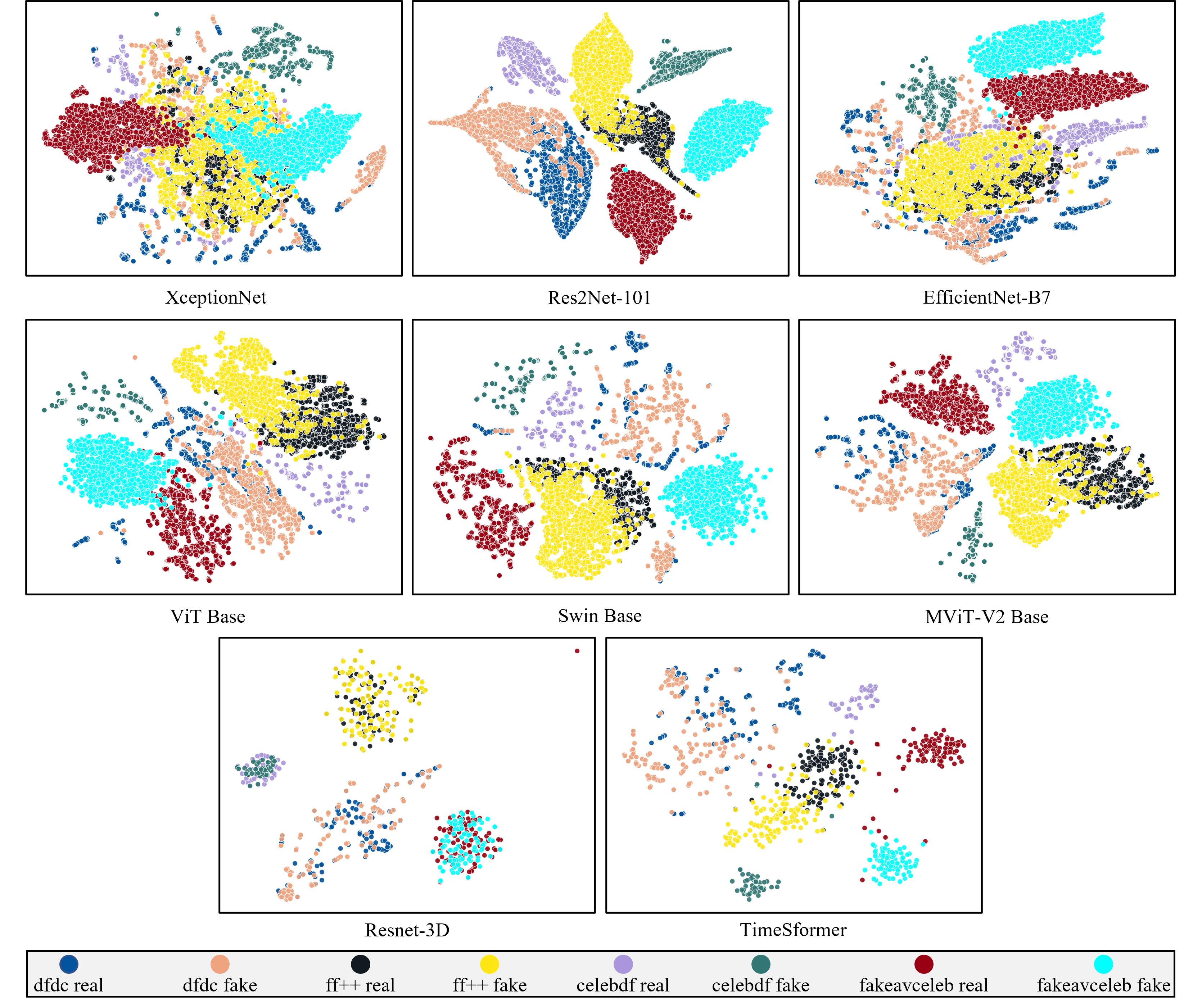}
\caption{TSNE visualisations of the participating detection models. We chose the best performing models on all datasets (with/without image augmentations).}
\label{fig:tsne}
\end{figure*}

Also, it can be noted that the transformer models (Swin-Base, and MViT-V2-Base) outperform the CNN based models. We can also see that the Res2Net-101 model also achieves excellent performance scores in intra-dataset evaluation setting, while having almost half the amount of parameters (43 million parameters) as compared to the best performing Swin-Base (87 million parameters) model. From Figure~\ref{fig:accuracy_comparison_all}, and Table~\ref{tab:allcomparisontable} we make one useful observation, i.e., models with multi-scale feature processing capabilities (Res2Net, MViT-V2, and Swin Transformer) are the best performing models out of all. 

Moving towards inter-dataset analysis, we illustrate results achieved by supervised models when evaluated in an inter-dataset setting in Figure~\ref{fig:interanalysis}. From the figure it is apparent that in inter-dataset evaluation, models perform in a significantly poor manner, as compared to intra-dataset evaluation. However, this is understandable i.e., detection models loose performance when tested on data coming from a different distribution. However, their is a useful finding in Figure~\ref{fig:interanalysis}, i.e., for all datasets, the best performing models are transformers.

In case of self-supervised models, by analysing the results in Table~\ref{tab:intra-selfsupervised} and \ref{tab:all-models-intra-self-supervised}, we can infer that self-supervised features (DINO) indeed provide better feature representations as compared to supervised feature representations. DINO seems to outperform both supervised ViT as well as CLIP as is also apparent from the ROC curvers illustrated in Figure~\ref{fig:selfsupervised_roc}.

We also visualise the TSNE plots for all the participating models in Figure~\ref{fig:tsne}, to get an idea about how the models separate real faces from those of the fake ones. Also, it gives us an idea about how the models group together faces coming from same datasets near to each other as compared to the faces coming from a different dataset. The TSNE plots also help us visualise which datasets are more challenging than the others. For example, if we look at the TSNE plots in Figure~\ref{fig:tsne}, we can see that the models tend to separate the easier datasets (FakeAVCeleb, and CelebDF-V2) in a better way, as compared to how they separate the more challenging datasets (FaceForensics++, and DFDC). We can see that the Res2Net-101 model does this separation in a better way as compared to the other models i.e., even better than the best performing Swin-Base model (we don't know how/why). The plots also support our findings that FaceForensics++ and DFDC are indeed challenging datasets, and that the models are not as good at separating the fake, and real faces coming from these datasets as they can separate the fake, and real faces coming from FakeAVCeleb and CelebDF-V2 datasets.

\begin{table}[h!]
  \centering
  \caption{This table compares the performance of all the participating (supervised) models. We present scores after averaging the scores (LogLoss, AUC, Accuracy) achieved by each model when evaluated in an intra-dataset setting as given in Equation~\ref{eq:Average_Score}.}
  \resizebox{.7\linewidth}{!}{%
    \begin{tabular}{l|c|c|c|c|c|c}
    \toprule
    \rowcolor[rgb]{ 0,  0,  0} \multicolumn{7}{c}{\textcolor[rgb]{ 1,  1,  1}{\textbf{Performance Comparison of All Models on All Datasets}}} \\
    \midrule
    \multicolumn{1}{c|}{\multirow{2}[3]{*}{\textbf{Model}}} & \multicolumn{3}{c|}{\textbf{With Augs}} & \multicolumn{3}{c}{\textbf{No Augs}} \\
\cmidrule{2-7}          & \textbf{LogLoss} & \textbf{AUC} & \textbf{ACC} & \textbf{LogLoss} & \textbf{AUC} & \textbf{ACC} \\
    \midrule
    Xception & 0.2179 & 96.36\% & 91.40\% & 0.2121 & 96.87\% & 92.02\% \\
    \midrule
    Res2Net-101 & \cellcolor[rgb]{ 1,  .753,  0}\textbf{0.1395} & 97.12\% & 93.10\% & 0.2282 & \cellcolor[rgb]{ 1,  .753,  0}\textbf{97.27\%} & 93.67\% \\
    \midrule
    EfficientNet-B7 & 0.2305 & 96.71\% & 91.92\% & 0.2097 & 96.83\% & 92.42\% \\
    \midrule
    ViT   & 0.2388 & 97.10\% & 91.62\% & 0.3350 & 95.55\% & 88.63\% \\
    \midrule
    Swin  & 0.1494 & \cellcolor[rgb]{ 1,  .753,  0}\textbf{97.35\%} & \cellcolor[rgb]{ 1,  .753,  0}\textbf{94.05\%} & 0.2275 & 96.34\% & 92.63\% \\
    \midrule
    MViT  & 0.1998 & 96.68\% & 93.16\% & 0.1882 & 97.07\% & \cellcolor[rgb]{ 1,  .753,  0}\textbf{93.76\%} \\
    \midrule
    Resnet-3D & 0.2470 & 95.44\% & 90.77\% & \cellcolor[rgb]{ 1,  .753,  0}\textbf{0.1620} & 94.89\% & 89.89\% \\
    \midrule
    TimeSformer & 0.2196 & 97.06\% & 92.18\% & 0.2521 & 97.24\% & 92.12\% \\
    \bottomrule
    \bottomrule
    \end{tabular}%
    }
  \label{tab:allcomparisontable}%
\end{table}%

Another finding which is quite apparent is that the image models do this separation job relatively better than those of the video models. This is understandable, as we pointed out above that the video models need relatively larger amounts of training data (in our case we train both the image/video models on the same amount of data). Also, the video models we use are not the newest, most powerful models out there, however, we choose these because of the lack of compute resources, and for the sack of experimentation.

We also present the ROC curves of the participating models evaluated in an intra-dataset setting in Figure~\ref{fig:roc}. The AUC scores achieved by the models also show that FakeAVCeleb and CelebDF-V2 datasets are easier to learn for the models, as compared to FaceForensics++ and DFDC datasets. This also suggests that while training the models for deepfake detection, it will give better generalisation capability to the models when they are trained on challenging datasets, rather than the easier ones.

% In Figure~\ref{fig:cbams} we visualise the CAMs (Class Activation Maps) of the supervised image models. CAMs indicate the discriminative regions of a given face which strongly influences the model to make a decision e.g., real or fake. For CNN models we extract CAMs from the last $conv$ layer. For ViT-Base we extract the maps from second last $attention-block$, whereas, in case of Swin-Base model we extract maps from the last $attention-block$.

\begin{equation}
\text{Score} = \frac{s_1 + s_2 + s_3 + s_4}{4}
\label{eq:Average_Score}
\end{equation}

where s1 refers to score (LogLoss, AUC, ACC) achieved by a model when trained and evaluated on dataset1, s2 refers to score (LogLoss, AUC, ACC) achieved by a model when trained and evaluated on dataset2 and so on. The scores reported in Tables~\ref{tab:allcomparisontable} and \ref{tab:intra-selfsupervised} for each model are calculated using this equation.

\begin{table}[htbp]
  \centering
  \caption{This table compares the performance of the self-supervised models. We present scores after averaging the scores (LogLoss, AUC, Accuracy) achieved by each model when evaluated in an intra-dataset setting, as given in Equation~\ref{eq:Average_Score}. In this table, \textit{\textbf{Supervised}} refer to ViT-Base model pre-trained using supervised training scheme. \textit{\textbf{DINO}} refers to ViT-Base model pre-trained using self-supervised scheme proposed in~\cite{Caron2021EmergingPI}, and \textit{\textbf{CLIP}} refers to ViT-Base pre-trained using self-supervised scheme prposed in~\cite{Radford2021LearningTV}. All of these ViT-Base models are used as feature extractors, where we only train a classification head on top of each of the feature extractor, and freeze the weights of feature extractors.}
  \resizebox{.65\linewidth}{!}{%
    \begin{tabular}{l|c|c|c|c|c|c}
    \toprule
    \rowcolor[rgb]{ 0,  0,  0} \multicolumn{7}{c}{\textcolor[rgb]{ 1,  1,  1}{\textbf{Performance Comparison of All Models on All Datasets}}} \\
    \midrule
    \multicolumn{1}{c|}{\multirow{2}[3]{*}{\textbf{Model}}} & \multicolumn{3}{c|}{\textbf{With Augs}} & \multicolumn{3}{c}{\textbf{No Augs}} \\
\cmidrule{2-7}          & \textbf{LogLoss} & \textbf{AUC} & \textbf{ACC} & \textbf{LogLoss} & \textbf{AUC} & \textbf{ACC} \\
    \midrule
    Supervised & \cellcolor[rgb]{ 1,  .753,  0}\textbf{0.4516} & 87.00\% & 78.54\% & \cellcolor[rgb]{ 1,  .753,  0}\textbf{0.4191} & 88.84\% & 81.59\% \\
    \midrule
    Dino  & 0.9924 & \cellcolor[rgb]{ 1,  .753,  0}\textbf{91.43\%} & \cellcolor[rgb]{ 1,  .753,  0}\textbf{84.80\%} & 0.7932 & \cellcolor[rgb]{ 1,  .753,  0}\textbf{92.31\%} & \cellcolor[rgb]{ 1,  .753,  0}\textbf{85.54\%} \\
    \midrule
    CLIP  & 1.0244 & 66.17\% & 62.26\% & 1.0513 & 69.40\% & 63.98\% \\
    \bottomrule
    \bottomrule
    \end{tabular}%
    }
  \label{tab:intra-selfsupervised}%
\end{table}%

\begin{table}[htbp]
  \centering
  \caption{This table compares the performance of all the participating (self-supervised) models when evaluated in an intra-dataset setting. The statistics of this table are illustrated in Figure~\ref{fig:selfsupervised_roc}.}
  \resizebox{.75\linewidth}{!}{%
    \begin{tabular}{l|c|c|c|c|c|c|c}
    \toprule
    \rowcolor[rgb]{ 0,  0,  0} \multicolumn{8}{c}{\textcolor[rgb]{ 1,  1,  1}{\textbf{Performance Comparison of All Models on Individual Datasets}}} \\
    \midrule
    \multicolumn{1}{c|}{\multirow{2}[3]{*}{\textbf{Model}}} & \multicolumn{3}{c|}{\textbf{With Augs}} & \multicolumn{3}{c|}{\textbf{No Augs}} & \multirow{2}[3]{*}{\textbf{Dataset}} \\
\cmidrule{2-7}          & \textbf{LogLoss} & \textbf{AUC} & \textbf{ACC} & \textbf{LogLoss} & \textbf{AUC} & \textbf{ACC} &  \\
    \midrule
    Supervised & 0.4105 & 90.19\% & 82.50\% & 0.3727 & 91.77\% & 85.50\% & \multirow{3}[6]{*}{\textbf{FakeAVCeleb}} \\
\cmidrule{1-7}    Dino  & \cellcolor[rgb]{ 1,  .753,  0}\textbf{0.1444} & \cellcolor[rgb]{ 1,  .753,  0}\textbf{99.00\%} & \cellcolor[rgb]{ 1,  .753,  0}\textbf{95.33\%} & \cellcolor[rgb]{ 1,  .753,  0}\textbf{0.0801} & \cellcolor[rgb]{ 1,  .753,  0}\textbf{99.64\%} & \cellcolor[rgb]{ 1,  .753,  0}\textbf{97.25\%} &  \\
\cmidrule{1-7}    CLIP  & 1.2851 & 63.93\% & 60.65\% & 1.3111 & 65.20\% & 61.13\% &  \\
    \midrule
    \rowcolor[rgb]{ .851,  .851,  .851}       &       &       &       &       &       &       &  \\
    \midrule
    Supervised & \cellcolor[rgb]{ 1,  .753,  0}\textbf{0.2941} & 95.52\% & 88.05\% & \cellcolor[rgb]{ 1,  .753,  0}\textbf{0.2237} & \cellcolor[rgb]{ 1,  .753,  0}\textbf{97.18\%} & \cellcolor[rgb]{ 1,  .753,  0}\textbf{91.80\%} & \multirow{3}[6]{*}{\textbf{CelebDF-V2}} \\
\cmidrule{1-7}    Dino  & 0.3655 & \cellcolor[rgb]{ 1,  .753,  0}\textbf{97.31\%} & \cellcolor[rgb]{ 1,  .753,  0}\textbf{90.90\%} & 0.3930 & 97.10\% & 88.90\% &  \\
\cmidrule{1-7}    CLIP  & 0.7811 & 77.00\% & 69.50\% & 0.6860 & 82.10\% & 74.10\% &  \\
    \midrule
    \rowcolor[rgb]{ .851,  .851,  .851}       &       &       &       &       &       &       &  \\
    \midrule
    Supervised & \cellcolor[rgb]{ 1,  .753,  0}\textbf{0.5182} & 83.11\% & 74.95\% & \cellcolor[rgb]{ 1,  .753,  0}\textbf{0.4971} & 85.47\% & 77.43\% & \multirow{3}[6]{*}{\textbf{FaceForensics++}} \\
\cmidrule{1-7}    Dino  & 1.1758 & \cellcolor[rgb]{ 1,  .753,  0}\textbf{88.67\%} & \cellcolor[rgb]{ 1,  .753,  0}\textbf{80.60\%} & 1.1186 & \cellcolor[rgb]{ 1,  .753,  0}\textbf{89.48\%} & \cellcolor[rgb]{ 1,  .753,  0}\textbf{81.85\%} &  \\
\cmidrule{1-7}    CLIP  & 0.9634 & 62.46\% & 59.05\% & 1.1070 & 62.61\% & 59.73\% &  \\
    \midrule
    \rowcolor[rgb]{ .851,  .851,  .851}       &       &       &       &       &       &       &  \\
    \midrule
    Supervised & \cellcolor[rgb]{ 1,  .753,  0}\textbf{0.5836} & 79.19\% & 68.65\% & \cellcolor[rgb]{ 1,  .753,  0}\textbf{0.5829} & 80.93\% & 72.63\% & \multirow{3}[6]{*}{\textbf{DFDC}} \\
\cmidrule{1-7}    Dino  & 2.2839 & \cellcolor[rgb]{ 1,  .753,  0}\textbf{80.72\%} & \cellcolor[rgb]{ 1,  .753,  0}\textbf{72.38\%} & 1.5812 & \cellcolor[rgb]{ 1,  .753,  0}\textbf{83.03\%} & \cellcolor[rgb]{ 1,  .753,  0}\textbf{74.15\%} &  \\
\cmidrule{1-7}    CLIP  & 1.0681 & 61.30\% & 59.83\% & 1.1011 & 67.69\% & 60.95\% &  \\
    \bottomrule
    \bottomrule
    \end{tabular}%
    }
  \label{tab:all-models-intra-self-supervised}%
\end{table}%
%%%%%%%%%%%%%%%%%%%%%%%%%%%%%%%%%%%%%%%%%%
\section{Conclusions}
\label{conclusion}

In conclusion, this paper investigates the performance of various image and video classification architectures (supervised, self-supervised) on the task of deepfake detection when trained and evaluated on four different datasets. We aimed at identifying which models perform better than other participating models, which model generalises well on unseen data as compared to the other models. Through experimentation and analysis of the achieved results we conclude that models possessing the capability of processing multi-scale features (Res2Net-101, MViT-V2, and SWIN Transformer) achieve better overall performance in intra-dataset comparison. For inter-dataset comparison, or in other words, the generalisation capability comparison, we infer from the results that transformer models perform better than that of the CNN models. It is also apparent from the results obtained through both inter-dataset as well as intra-dataset comparisons, the image augmentations do not always help achieve better performance scores.

Through intra-dataset comparisons we establish that the DFDC dataset is the most challenging dataset to learn for the models, whereas FaceForensics++ dataset is ranked as second most challenging dataset. However, through inter-dataset evaluation, we establish that the FaceForensics++ dataset offers the best generalisation capabilities to the models, as compared to other datasets. DFDC ranks second in providing the generalisation capabilities. The remaining two datasets FakeAVCeleb, and CelebDF-V2 appear to be fairly easy for the models to learn and achieve excellent performance in intra-dataset comparison. However, they do not provide the models with any generalisation capability, i.e., models trained using these datasets perform poorly when evaluated on other datasets.

In addition to analysing supervised image/video recognition models, we also explore the performance of self-supervised models for deepfake detection in an intra-dataset setting. Through our experiments we find the ViT-Base model which was pre-trained using DINO~\cite{Caron2021EmergingPI} to achieve better performance as compared to the supervised ViT-Base, and self-supervised CLIP ViT-Base. We also find in these experiments these models achieve better performance scores when trained without using image augmentations.

% Overall, the findings of this study provide insights into (1) the effectiveness of different models for deepfake detection, (2) the datasets which are more capable of providing the generalisation capabilities to the modes, and (3) whether self-supervised features provide better representations for deepfake image classification. 

All in all, we present a detailed analysis of the performance achieved by several different deepfake detection architectures on four different deepfake detection benchmarks. We carry out extensive experiments and provide detailed results along with useful visualisations to help understand the overall contributions of this paper to the reader. We regard this study as an entry point for researchers exploring the research field of deepfake detection who are trying to make sense of different architectures and datasets to develop their own solutions. We are confident that this study provides useful insights into the problem of deepfake detection. 

% Early-stage researchers can make informed decisions at the start of their research in this field e.g., it is now easier to choose suitable architectures for deepfake detection, as well as which dataset can be more beneficial. 

In future work, we aim at analysing even more diverse set of architectures, and newer datasets. In addition to that we plan to focus more towards self-supervised training strategies to train models, as well as try to incorporate knowledge distillation, domain adaptation strategies to help make models better at classifying unseen samples correctly.

% A TABLE RANKING ALL THE MODELS BASED ON BEST PERFORMANCES ON EACH DATASET BOTH ON INTER AS WELL AS INTRA EVALUATION SETTINGS.

% A table or figure showing which dataset is more challenging to learn, and which provides best generalization.

%\clearpage
\begin{acks}
This research was supported by industry partners and the Research Council of Norway with funding to MediaFutures: Research Centre for Responsible Media Technology and Innovation, through the Centres for Research-based Innovation scheme, project number 309339.
\end{acks}

\bibliographystyle{ACM-Reference-Format}
\bibliography{main}

\appendix
\section{Inter-Dataset Evaluation Scores}

\begin{figure*}[!htb]
\centering
\includegraphics[width=0.95\linewidth]{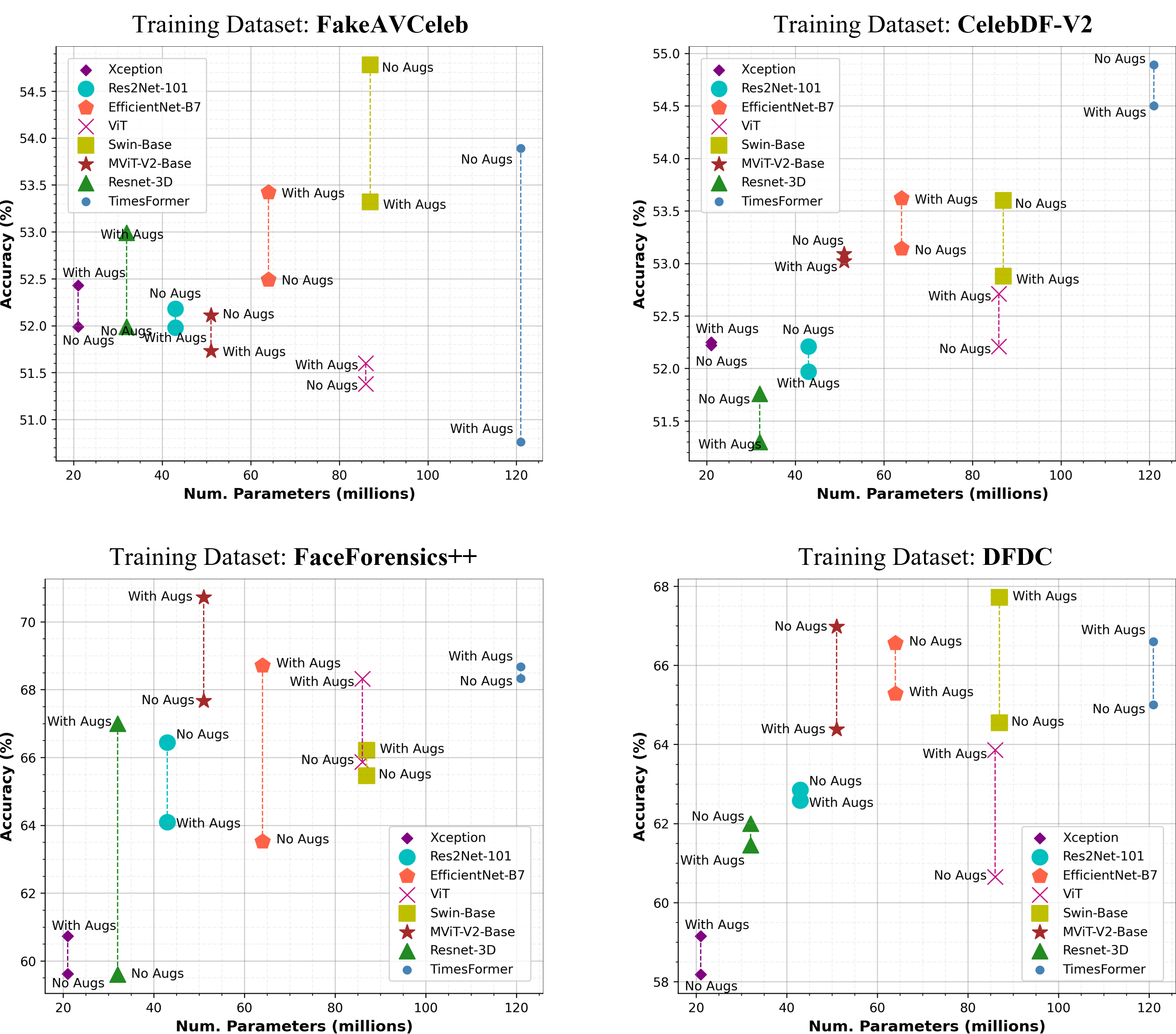}
\caption{Performance (accuracy) comparison of participating models evaluated using inter-dataset scheme. Results in this figure are obtained by, (1) evaluating each model trained on one dataset on each of the remaining datasets, and (2) averaging the achieved scores, i.e., add the 3 accuracy scores and divide by 3.}
\label{fig:interanalysis}
\end{figure*}

\begin{table}[htbp]
  \centering
  \tiny
  \caption{This table compares the performance of all the participating (supervised) models evaluated in an inter-dataset setting. We present scores after averaging the scores (LogLoss, AUC, Accuracy) achieved by each model on each of the datasets. Figure~\ref{fig:interanalysis} illustrate the statistics of this table.}
  \resizebox{0.85\linewidth}{!}{%
    \begin{tabular}{l|c|c|c|c|c|c|c}
    \toprule
    \rowcolor[rgb]{ 0,  0,  0} \multicolumn{7}{c|}{\textcolor[rgb]{ 1,  1,  1}{\textbf{Inter-Dataset Evaluation}}} & \multicolumn{1}{c}{\multirow{3}[6]{*}{\cellcolor[rgb]{ 1,  1,  1}\textbf{Training Dataset}}} \\
\cmidrule{1-7}    \multicolumn{1}{c|}{\multirow{2}[4]{*}{\textbf{Model}}} & \multicolumn{3}{c|}{\textbf{With Augs}} & \multicolumn{3}{c|}{\textbf{No Augs}} &  \\
\cmidrule{2-7}          & \textbf{LogLoss} & \textbf{AUC} & \textbf{ACC} & \textbf{LogLoss} & \textbf{AUC} & \textbf{ACC} &  \\
    \midrule
    Xception & 7.4484 & 57.69\% & 52.43\% & 6.8728 & 55.41\% & 51.99\% & \multirow{8}[15]{*}{\textbf{FakeAVCeleb}} \\
\cmidrule{1-7}    Res2Net-101 & 8.5574 & 58.77\% & 51.98\% & 7.0666 & 60.64\% & 52.18\% &  \\
\cmidrule{1-7}    EfficientNet-B7 & 8.5664 & 62.32\% & \cellcolor[rgb]{ 1,  .753,  0}\textbf{53.42\%} & 10.7718 & 60.05\% & 52.49\% &  \\
\cmidrule{1-7}    ViT   & 6.7348 & 61.01\% & 51.60\% & 9.1672 & 58.45\% & 51.38\% &  \\
\cmidrule{1-7}    Swin  & 5.1077 & \cellcolor[rgb]{ 1,  .753,  0}\textbf{62.54\%} & 53.32\% & 4.3274 & \cellcolor[rgb]{ 1,  .753,  0}\textbf{64.88\%} & \cellcolor[rgb]{ 1,  .753,  0}\textbf{54.78\%} &  \\
\cmidrule{1-7}    MViT  & 4.7564 & 58.78\% & 51.73\% & 4.2891 & 59.38\% & 52.11\% &  \\
\cmidrule{1-7}    ResNet-3D & \cellcolor[rgb]{ 1,  .753,  0}\textbf{4.4308} & 57.61\% & 52.99\% & \cellcolor[rgb]{ 1,  .753,  0}\textbf{3.8206} & 60.09\% & 51.99\% &  \\
\cmidrule{1-7}    TimeSformer & 4.7334 & 61.55\% & 50.76\% & 4.7759 & 63.95\% & 53.89\% &  \\
    \rowcolor[rgb]{ .816,  .808,  .808} \multicolumn{1}{c}{} & \multicolumn{1}{c}{} & \multicolumn{1}{c}{} & \multicolumn{1}{c}{} & \multicolumn{1}{c}{} & \multicolumn{1}{c}{} & \multicolumn{1}{c}{} &  \\
    \midrule
    Xception & \cellcolor[rgb]{ 1,  .753,  0}\textbf{3.9439} & 65.06\% & 52.25\% & 4.8776 & 66.40\% & 52.22\% & \multirow{8}[16]{*}{\textbf{CelebDF-V2}} \\
\cmidrule{1-7}    Res2Net-101 & 5.4266 & 65.90\% & 51.97\% & 5.6891 & 66.21\% & 52.21\% &  \\
\cmidrule{1-7}    EfficientNet-B7 & 5.9514 & 66.99\% & 53.62\% & 8.9668 & 67.13\% & 53.14\% &  \\
\cmidrule{1-7}    ViT   & 5.4921 & 68.52\% & 52.71\% & 8.9981 & 66.36\% & 52.21\% &  \\
\cmidrule{1-7}    Swin  & 5.6007 & 70.06\% & 52.88\% & 4.8405 & \cellcolor[rgb]{ 1,  .753,  0}\textbf{70.56\%} & 53.60\% &  \\
\cmidrule{1-7}    MViT  & 4.8723 & \cellcolor[rgb]{ 1,  .753,  0}\textbf{70.71\%} & 53.02\% & \cellcolor[rgb]{ 1,  .753,  0}\textbf{4.6419} & 67.20\% & 53.09\% &  \\
\cmidrule{1-7}    ResNet-3D & 6.8365 & 61.57\% & 51.30\% & 5.0504 & 64.52\% & 51.76\% &  \\
\cmidrule{1-7}    TimeSformer & 4.5629 & 69.04\% & \cellcolor[rgb]{ 1,  .753,  0}\textbf{54.50\%} & 4.9391 & 69.43\% & \cellcolor[rgb]{ 1,  .753,  0}\textbf{54.89\%} &  \\
    \midrule
    \rowcolor[rgb]{ .816,  .808,  .808} \multicolumn{1}{c}{} & \multicolumn{1}{c}{} & \multicolumn{1}{c}{} & \multicolumn{1}{c}{} & \multicolumn{1}{c}{} & \multicolumn{1}{c}{} & \multicolumn{1}{c}{} &  \\
    \midrule
    Xception & 1.0701 & 69.92\% & 60.73\% & 1.1262 & 67.78\% & 59.62\% & \multirow{8}[16]{*}{\textbf{FaceForensics++}} \\
\cmidrule{1-7}    Res2Net-101 & 1.0165 & 73.46\% & 64.09\% & 1.2360 & 73.61\% & 66.44\% &  \\
\cmidrule{1-7}    EfficientNet-B7 & 0.8792 & 79.51\% & 68.71\% & 1.0068 & 69.80\% & 63.52\% &  \\
\cmidrule{1-7}    ViT   & \cellcolor[rgb]{ 1,  .753,  0}\textbf{0.7899} & 78.45\% & 68.32\% & 0.8301 & 73.24\% & 65.87\% &  \\
\cmidrule{1-7}    Swin  & 0.8517 & 77.94\% & 66.21\% & 0.8482 & \cellcolor[rgb]{ 1,  .753,  0}\textbf{78.03\%} & 65.46\% &  \\
\cmidrule{1-7}    MViT  & 0.8407 & \cellcolor[rgb]{ 1,  .753,  0}\textbf{79.75\%} & \cellcolor[rgb]{ 1,  .753,  0}\textbf{70.72\%} & \cellcolor[rgb]{ 1,  .753,  0}\textbf{0.7292} & 75.85\% & 67.67\% &  \\
\cmidrule{1-7}    ResNet-3D & 1.0639 & 74.47\% & 67.00\% & 1.3331 & 66.61\% & 59.50\% &  \\
\cmidrule{1-7}    TimeSformer & 1.0665 & 75.59\% & 68.67\% & 0.8492 & 77.03\% & \cellcolor[rgb]{ 1,  .753,  0}\textbf{68.33\%} &  \\
    \midrule
    \rowcolor[rgb]{ .816,  .808,  .808} \multicolumn{1}{c}{} & \multicolumn{1}{c}{} & \multicolumn{1}{c}{} & \multicolumn{1}{c}{} & \multicolumn{1}{c}{} & \multicolumn{1}{c}{} & \multicolumn{1}{c}{} &  \\
    \midrule
    Xception & 1.2959 & 63.62\% & 59.15\% & 1.6780 & 64.18\% & 58.18\% & \multirow{8}[16]{*}{\textbf{DFDC}} \\
\cmidrule{1-7}    Res2Net-101 & 2.0224 & 67.80\% & 62.58\% & 1.7396 & 69.50\% & 62.85\% &  \\
\cmidrule{1-7}    EfficientNet-B7 & \cellcolor[rgb]{ 1,  .753,  0}\textbf{1.0388} & 71.32\% & 65.28\% & 1.2764 & 72.28\% & 66.56\% &  \\
\cmidrule{1-7}    ViT   & 1.2198 & 70.45\% & 63.86\% & 1.2498 & 64.71\% & 60.65\% &  \\
\cmidrule{1-7}    Swin  & 1.2423 & \cellcolor[rgb]{ 1,  .753,  0}\textbf{73.49\%} & \cellcolor[rgb]{ 1,  .753,  0}\textbf{67.72\%} & 1.3802 & 69.49\% & 64.55\% &  \\
\cmidrule{1-7}    MViT  & 1.2329 & 72.37\% & 64.38\% & 1.2254 & \cellcolor[rgb]{ 1,  .753,  0}\textbf{72.68\%} & \cellcolor[rgb]{ 1,  .753,  0}\textbf{66.98\%} &  \\
\cmidrule{1-7}    ResNet-3D & 1.1354 & 66.69\% & 61.45\% & \cellcolor[rgb]{ 1,  .753,  0}\textbf{1.1354} & 66.27\% & 62.00\% &  \\
\cmidrule{1-7}    TimeSformer & 1.1421 & 70.66\% & 66.60\% & 1.6584 & 71.77\% & 65.00\% &  \\
    \bottomrule
    \bottomrule
    \end{tabular}%
    }
  \label{tab:allcomparisoninterdataset}%
\end{table}%

\begin{table}[htbp]
  \centering
  \caption{Inter-dataset evaluation scores of models trained on FakeAVCeleb~\cite{Khalid2021FakeAVCelebAN} dataset and evaluated on the remaining three datasets.}
  \resizebox{0.5\linewidth}{!}{%
    \begin{tabular}{l|c|c|c|c|c|c|c}
    \toprule
    \rowcolor[rgb]{ 0,  0,  0} \multicolumn{7}{c|}{\textcolor[rgb]{ 1,  1,  1}{\textbf{Training Dataset: FakeAVCeleb}}} & \multicolumn{1}{c}{\multirow{2}[6]{*}{\cellcolor[rgb]{ 1,  1,  1}\textbf{Evaluation Dataset}}} \\
\cmidrule{1-7}    \multicolumn{1}{c|}{\multirow{2}[4]{*}{\textbf{Model}}} & \multicolumn{3}{c|}{\textbf{With Augs}} & \multicolumn{3}{c|}{\textbf{No Augs}} &  \\
\cmidrule{2-7}          & \textbf{LogLoss} & \textbf{AUC} & \textbf{ACC} & \textbf{LogLoss} & \textbf{AUC} & \textbf{ACC} &  \\
    \midrule
    Xception & 3.1366 & 50.52\% & 56.10\% & 3.9659 & 44.63\% & 54.25\% & \multirow{6}[15]{*}{\textbf{CelebDF-V2}} \\
\cmidrule{1-7}    Res2Net-101 & 3.9007 & 57.61\% & 54.60\% & 3.2127 & 56.73\% & 54.50\% &  \\
\cmidrule{1-7}    EfficientNet-B7 & 5.7925 & 59.31\% & 54.50\% & 12.3786 & 54.30\% & 51.65\% &  \\
\cmidrule{1-7}    ViT   & 4.7035 & 56.51\% & 51.60\% & 7.0900 & 52.18\% & 46.65\% &  \\
\cmidrule{1-7}    Swin  & 3.5148 & 61.13\% & \cellcolor[rgb]{ 1,  .753,  0}\textbf{57.70\%} & 2.8360 & \cellcolor[rgb]{ 1,  .753,  0}\textbf{62.54\%} & \cellcolor[rgb]{ 1,  .753,  0}\textbf{61.45\%} &  \\
\cmidrule{1-7}    MViT  & 4.5526 & \cellcolor[rgb]{ 1,  .753,  0}\textbf{65.37\%} & 54.00\% & 3.6492 & 58.18\% & 55.05\% &  \\
\cmidrule{1-7}    ResNet-3D & \cellcolor[rgb]{ 1,  .753,  0}\textbf{2.4752} & 53.88\% & 51.00\% & \cellcolor[rgb]{ 1,  .753,  0}\textbf{1.7475} & 57.36\% & 49.00\% &  \\
\cmidrule{1-7}    TimeSformer & 3.9086 & 51.60\% & 48.00\% & 3.2374 & 58.52\% & 55.00\% &  \\
\midrule
    \rowcolor[rgb]{ .816,  .808,  .808} \multicolumn{1}{c}{} & \multicolumn{1}{c}{} & \multicolumn{1}{c}{} & \multicolumn{1}{c}{} & \multicolumn{1}{c}{} & \multicolumn{1}{c}{} & \multicolumn{1}{c}{} &  \\
    \midrule
    Xception & 10.3539 & \cellcolor[rgb]{ 1,  .753,  0}\textbf{62.90\%} & 50.38\% & 8.9711 & 63.06\% & 50.33\% & \multirow{6}[16]{*}{\textbf{FaceForensics++}} \\
\cmidrule{1-7}    Res2Net-101 & 11.6456 & 59.23\% & 50.18\% & 9.7934 & 58.54\% & 50.53\% &  \\
\cmidrule{1-7}    EfficientNet-B7 & 10.5412 & 63.80\% & \cellcolor[rgb]{ 1,  .753,  0}\textbf{52.45\%} & 10.4825 & 63.13\% & \cellcolor[rgb]{ 1,  .753,  0}\textbf{52.05\%} &  \\
\cmidrule{1-7}    ViT   & 9.3036 & 61.70\% & 51.10\% & 12.3768 & 58.44\% & 50.93\% &  \\
\cmidrule{1-7}    Swin  & 6.1675 & 62.97\% & 50.70\% & 5.5166 & \cellcolor[rgb]{ 1,  .753,  0}\textbf{64.87\%} & 51.23\% &  \\
\cmidrule{1-7}    MViT  & \cellcolor[rgb]{ 1,  .753,  0}\textbf{4.8833} & 56.51\% & 50.65\% & \cellcolor[rgb]{ 1,  .753,  0}\textbf{4.7116} & 63.40\% & 50.85\% &  \\
\cmidrule{1-7}    ResNet-3D & 6.7343 & 51.30\% & 50.71\% & 5.9796 & 53.99\% & 50.71\% &  \\
\cmidrule{1-7}    TimeSformer & 6.0010 & 62.61\% & 51.79\% & 6.1889 & 62.60\% & 51.43\% &  \\
    \midrule
    \rowcolor[rgb]{ .816,  .808,  .808} \multicolumn{1}{c}{} & \multicolumn{1}{c}{} & \multicolumn{1}{c}{} & \multicolumn{1}{c}{} & \multicolumn{1}{c}{} & \multicolumn{1}{c}{} & \multicolumn{1}{c}{} &  \\
    \midrule
    Xception & 8.8546 & 59.65\% & 50.80\% & 7.6813 & 58.53\% & 51.40\% & \multirow{6}[16]{*}{\textbf{DFDC}} \\
\cmidrule{1-7}    Res2Net-101 & 10.1260 & 59.48\% & 51.15\% & 8.1937 & 66.66\% & 51.50\% &  \\
\cmidrule{1-7}    EfficientNet-B7 & 9.3656 & 63.86\% & 53.30\% & 9.4543 & 62.71\% & 53.78\% &  \\
\cmidrule{1-7}    ViT   & 6.1972 & 64.81\% & 52.10\% & 8.0348 & 64.71\% & 56.58\% &  \\
\cmidrule{1-7}    Swin  & 5.6410 & 63.51\% & 51.55\% & 4.6297 & 67.25\% & 51.65\% &  \\
\cmidrule{1-7}    MViT  & 4.8333 & 54.46\% & 50.55\% & 4.5065 & 56.55\% & 50.43\% &  \\
\cmidrule{1-7}    ResNet-3D & \cellcolor[rgb]{ 1,  .753,  0}\textbf{4.0828} & 67.65\% & \cellcolor[rgb]{ 1,  .753,  0}\textbf{57.25\%} & \cellcolor[rgb]{ 1,  .753,  0}\textbf{3.7347} & 68.91\% & \cellcolor[rgb]{ 1,  .753,  0}\textbf{56.25\%} &  \\
\cmidrule{1-7}    TimeSformer & 4.2907 & \cellcolor[rgb]{ 1,  .753,  0}\textbf{70.43\%} & 52.50\% & 4.9015 & \cellcolor[rgb]{ 1,  .753,  0}\textbf{70.74\%} & 55.25\% &  \\
    \bottomrule
    \bottomrule
    \end{tabular}%
    }
  \label{tab:inter-fakeavceleb}%
\end{table}%

\begin{table}[htbp]
  \centering
  \caption{Inter-dataset evaluation scores of models trained on CelebDF-V2~\cite{Li2020CelebDFAL} dataset and evaluated on the remaining three datasets.}
  \resizebox{.5\linewidth}{!}{%
    \begin{tabular}{l|c|c|c|c|c|c|c}
    \toprule
    \rowcolor[rgb]{ 0,  0,  0} \multicolumn{7}{c|}{\textcolor[rgb]{ 1,  1,  1}{\textbf{Training Dataset: CelebDF-V2}}} & \multicolumn{1}{c}{\multirow{2}[6]{*}{\cellcolor[rgb]{ 1,  1,  1}\textbf{Evaluation Dataset}}} \\
\cmidrule{1-7}    \multicolumn{1}{c|}{\multirow{2}[4]{*}{\textbf{Model}}} & \multicolumn{3}{c|}{\textbf{With Augs}} & \multicolumn{3}{c|}{\textbf{No Augs}} &  \\
\cmidrule{2-7}          & \textbf{LogLoss} & \textbf{AUC} & \textbf{ACC} & \textbf{LogLoss} & \textbf{AUC} & \textbf{ACC} &  \\
    \midrule
    Xception & \cellcolor[rgb]{ 1,  .753,  0}\textbf{4.7313} & 65.82\% & 51.68\% & 5.1136 & 67.77\% & 51.78\% & \multirow{6}[15]{*}{\textbf{FakeAVCeleb}} \\
\cmidrule{1-7}    Res2Net-101 & 5.7429 & 69.08\% & 52.30\% & \cellcolor[rgb]{ 1,  .753,  0}\textbf{4.3332} & \cellcolor[rgb]{ 1,  .753,  0}\textbf{71.01\%} & 52.83\% &  \\
\cmidrule{1-7}    EfficientNet-B7 & 7.4940 & 63.86\% & 52.05\% & 9.6846 & 65.93\% & 51.45\% &  \\
\cmidrule{1-7}    ViT   & 5.0347 & 69.00\% & \cellcolor[rgb]{ 1,  .753,  0}\textbf{53.08\%} & 9.6735 & 61.89\% & 52.18\% &  \\
\cmidrule{1-7}    Swin  & 6.0084 & 67.63\% & 52.20\% & 4.8922 & 68.28\% & 52.70\% &  \\
\cmidrule{1-7}    MViT  & 5.1980 & \cellcolor[rgb]{ 1,  .753,  0}\textbf{72.43\%} & 52.75\% & 5.3953 & 61.24\% & 51.55\% &  \\
\cmidrule{1-7}    ResNet-3D & 6.0756 & 62.79\% & 50.50\% & 4.9703 & 61.58\% & 50.50\% &  \\
\cmidrule{1-7}    TimeSformer & 4.8465 & 69.73\% & 53.00\% & 5.8829 & 68.77\% & \cellcolor[rgb]{ 1,  .753,  0}\textbf{54.00\%} &  \\
\midrule
    \rowcolor[rgb]{ .851,  .851,  .851} \multicolumn{1}{c}{} & \multicolumn{1}{c}{} & \multicolumn{1}{c}{} & \multicolumn{1}{c}{} & \multicolumn{1}{c}{} & \multicolumn{1}{c}{} & \multicolumn{1}{c}{} &  \\
    \midrule
    Xception & \cellcolor[rgb]{ 1,  .753,  0}\textbf{4.2473} & 63.26\% & 53.53\% & 5.6357 & 63.68\% & 53.58\% & \multirow{6}[16]{*}{\textbf{FaceForensics++}} \\
\cmidrule{1-7}    Res2Net-101 & 6.3947 & 64.79\% & 53.33\% & 6.9000 & 63.59\% & 52.90\% &  \\
\cmidrule{1-7}    EfficientNet-B7 & 6.3164 & 65.07\% & 54.80\% & 8.9065 & 66.31\% & 53.98\% &  \\
\cmidrule{1-7}    ViT   & 6.1010 & 68.14\% & 53.53\% & 9.9676 & 65.50\% & 53.50\% &  \\
\cmidrule{1-7}    Swin  & 6.0278 & \cellcolor[rgb]{ 1,  .753,  0}\textbf{70.13\%} & 54.23\% & 5.5408 & \cellcolor[rgb]{ 1,  .753,  0}\textbf{68.45\%} & 54.30\% &  \\
\cmidrule{1-7}    MViT  & 5.2175 & 70.01\% & 53.15\% & \cellcolor[rgb]{ 1,  .753,  0}\textbf{4.6160} & 67.88\% & 54.08\% &  \\
\cmidrule{1-7}    ResNet-3D & 7.1877 & 60.00\% & 52.14\% & 5.6544 & 66.01\% & 54.29\% &  \\
\cmidrule{1-7}    TimeSformer & 4.8228 & 68.84\% & \cellcolor[rgb]{ 1,  .753,  0}\textbf{57.50\%} & 5.0219 & 67.55\% & \cellcolor[rgb]{ 1,  .753,  0}\textbf{56.43\%} &  \\
    \midrule
    \rowcolor[rgb]{ .851,  .851,  .851} \multicolumn{1}{c}{} & \multicolumn{1}{c}{} & \multicolumn{1}{c}{} & \multicolumn{1}{c}{} & \multicolumn{1}{c}{} & \multicolumn{1}{c}{} & \multicolumn{1}{c}{} &  \\
    \midrule
    Xception & \cellcolor[rgb]{ 1,  .753,  0}\textbf{2.8532} & 66.11\% & 51.55\% & \cellcolor[rgb]{ 1,  .753,  0}\textbf{3.8835} & 67.74\% & 51.30\% & \multirow{6}[16]{*}{\textbf{DFDC}} \\
\cmidrule{1-7}    Res2Net-101 & 4.1424 & 63.83\% & 50.28\% & 5.8342 & 64.01\% & 50.90\% &  \\
\cmidrule{1-7}    EfficientNet-B7 & 4.0438 & 72.05\% & \cellcolor[rgb]{ 1,  .753,  0}\textbf{54.00\%} & 8.3092 & 69.17\% & 54.00\% &  \\
\cmidrule{1-7}    ViT   & 5.3405 & 68.41\% & 51.53\% & 7.3534 & 71.69\% & 50.95\% &  \\
\cmidrule{1-7}    Swin  & 4.7659 & \cellcolor[rgb]{ 1,  .753,  0}\textbf{72.42\%} & 52.20\% & 4.0886 & \cellcolor[rgb]{ 1,  .753,  0}\textbf{74.95\%} & 53.80\% &  \\
\cmidrule{1-7}    MViT  & 4.2014 & 69.69\% & 53.15\% & 3.9144 & 72.48\% & 53.65\% &  \\
\cmidrule{1-7}    ResNet-3D & 7.2461 & 61.91\% & 51.25\% & 4.5265 & 65.97\% & 50.50\% &  \\
\cmidrule{1-7}    TimeSformer & 4.0195 & 68.56\% & 53.00\% & 3.9124 & 71.98\% & \cellcolor[rgb]{ 1,  .753,  0}\textbf{54.25\%} &  \\
    \bottomrule
    \bottomrule
    \end{tabular}%
    }
  \label{tab:inter-celeb}%
\end{table}%

\begin{table}[htbp]
  \centering
  \caption{Inter-dataset evaluation scores of models trained on FaceForensics++~\cite{Rssler2019FaceForensicsLT} dataset and evaluated on the remaining three datasets.}
  \resizebox{.5\linewidth}{!}{%
    \begin{tabular}{l|c|c|c|c|c|c|c}
    \toprule
    \rowcolor[rgb]{ 0,  0,  0} \multicolumn{7}{c|}{\textcolor[rgb]{ 1,  1,  1}{\textbf{Training Dataset: FaceForensics++}}} & \multicolumn{1}{c}{\multirow{2}[6]{*}{\cellcolor[rgb]{ 1,  1,  1}\textbf{Evaluation Dataset}}} \\
\cmidrule{1-7}    \multicolumn{1}{c|}{\multirow{2}[4]{*}{\textbf{Model}}} & \multicolumn{3}{c|}{\textbf{With Augs}} & \multicolumn{3}{c|}{\textbf{No Augs}} &  \\
\cmidrule{2-7}          & \textbf{LogLoss} & \textbf{AUC} & \textbf{ACC} & \textbf{LogLoss} & \textbf{AUC} & \textbf{ACC} &  \\
    \midrule
    Xception & 0.8691 & 79.62\% & 65.88\% & 0.7795 & 76.14\% & 66.93\% & \multirow{6}[15]{*}{\textbf{FakeAVCeleb}} \\
\cmidrule{1-7}    Res2Net-101 & 0.7693 & 83.01\% & 71.28\% & 0.6527 & 85.48\% & 76.83\% &  \\
\cmidrule{1-7}    EfficientNet-B7 & 0.5782 & 89.59\% & 77.05\% & 0.7375 & 77.88\% & 70.08\% &  \\
\cmidrule{1-7}    ViT   & 0.6648 & 83.05\% & 70.65\% & 0.7419 & 76.40\% & 69.23\% &  \\
\cmidrule{1-7}    Swin  & 0.5880 & 87.72\% & 72.95\% & 0.6373 & 89.10\% & 71.15\% &  \\
\cmidrule{1-7}    MViT  & \cellcolor[rgb]{ 1,  .753,  0}\textbf{0.3654} & \cellcolor[rgb]{ 1,  .753,  0}\textbf{92.96\%} & \cellcolor[rgb]{ 1,  .753,  0}\textbf{84.65\%} & \cellcolor[rgb]{ 1,  .753,  0}\textbf{0.4047} & \cellcolor[rgb]{ 1,  .753,  0}\textbf{90.25\%} & \cellcolor[rgb]{ 1,  .753,  0}\textbf{81.90\%} &  \\
\cmidrule{1-7}    ResNet-3D & 0.7903 & 83.55\% & 68.00\% & 1.1338 & 73.34\% & 62.50\% &  \\
\cmidrule{1-7}    TimeSformer & 0.9135 & 79.33\% & 75.00\% & 0.7900 & 76.65\% & 70.50\% &  \\ 
\midrule
    \rowcolor[rgb]{ .851,  .851,  .851} \multicolumn{1}{c}{} & \multicolumn{1}{c}{} & \multicolumn{1}{c}{} & \multicolumn{1}{c}{} & \multicolumn{1}{c}{} & \multicolumn{1}{c}{} & \multicolumn{1}{c}{} &  \\
    \midrule
    Xception & 1.0426 & 65.92\% & 61.60\% & 1.2566 & 62.39\% & 58.65\% & \multirow{6}[16]{*}{\textbf{CelebDF-V2}} \\
\cmidrule{1-7}    Res2Net-101 & 1.0751 & 67.85\% & 62.40\% & 1.4218 & 65.46\% & 59.80\% &  \\
\cmidrule{1-7}    EfficientNet-B7 & 0.7759 & 78.46\% & 69.95\% & 1.0103 & 67.24\% & 61.25\% &  \\
\cmidrule{1-7}    ViT   & \cellcolor[rgb]{ 1,  .753,  0}\textbf{0.5915} & \cellcolor[rgb]{ 1,  .753,  0}\textbf{82.44\%} & \cellcolor[rgb]{ 1,  .753,  0}\textbf{74.10\%} & 0.8504 & 75.11\% & 65.40\% &  \\
\cmidrule{1-7}    Swin  & 0.7136 & 74.58\% & 67.05\% & 0.7879 & 70.94\% & 63.75\% &  \\
\cmidrule{1-7}    MViT  & 0.9791 & 76.66\% & 65.35\% & 0.7912 & 68.69\% & 62.70\% &  \\
\cmidrule{1-7}    ResNet-3D & 1.1992 & 66.12\% & 65.00\% & 1.5866 & 59.44\% & 55.00\% &  \\
\cmidrule{1-7}    TimeSformer & 1.1745 & 73.68\% & 63.00\% & \cellcolor[rgb]{ 1,  .753,  0}\textbf{0.7446} & \cellcolor[rgb]{ 1,  .753,  0}\textbf{80.40\%} & \cellcolor[rgb]{ 1,  .753,  0}\textbf{71.00\%} &  \\
    \midrule
    \rowcolor[rgb]{ .851,  .851,  .851} \multicolumn{1}{c}{} & \multicolumn{1}{c}{} & \multicolumn{1}{c}{} & \multicolumn{1}{c}{} & \multicolumn{1}{c}{} & \multicolumn{1}{c}{} & \multicolumn{1}{c}{} &  \\
    \midrule
    Xception & 1.2988 & 64.22\% & 54.70\% & 1.3424 & 64.81\% & 53.28\% & \multirow{6}[16]{*}{\textbf{DFDC}} \\
\cmidrule{1-7}    Res2Net-101 & 1.2052 & 69.51\% & 58.60\% & 1.6336 & 69.89\% & 62.70\% &  \\
\cmidrule{1-7}    EfficientNet-B7 & 1.2835 & 70.49\% & 59.13\% & 1.2726 & 64.29\% & 59.23\% &  \\
\cmidrule{1-7}    ViT   & 1.1135 & 69.87\% & 60.20\% & \cellcolor[rgb]{ 1,  .753,  0}\textbf{0.8981} & 68.20\% & 62.98\% &  \\
\cmidrule{1-7}    Swin  & 1.2534 & 71.53\% & 58.63\% & 1.1194 & \cellcolor[rgb]{ 1,  .753,  0}\textbf{74.04\%} & 61.48\% &  \\
\cmidrule{1-7}    MViT  & 1.1775 & 69.63\% & 62.15\% & 0.9917 & 68.61\% & 58.40\% &  \\
\cmidrule{1-7}    ResNet-3D & 1.2023 & 73.75\% & \cellcolor[rgb]{ 1,  .753,  0}\textbf{68.00\%} & 1.2788 & 67.04\% & 61.00\% &  \\
\cmidrule{1-7}    TimeSformer & \cellcolor[rgb]{ 1,  .753,  0}\textbf{1.1116} & \cellcolor[rgb]{ 1,  .753,  0}\textbf{73.77\%} & \cellcolor[rgb]{ 1,  .753,  0}\textbf{68.00\%} & 1.0129 & \cellcolor[rgb]{ 1,  .753,  0}\textbf{74.04\%} & \cellcolor[rgb]{ 1,  .753,  0}\textbf{63.50\%} &  \\
    \bottomrule
    \bottomrule
    \end{tabular}%
    }
  \label{tab:inter-ff}%
\end{table}%

\begin{table}[htbp]
  \centering
  \caption{Inter-dataset evaluation scores of models trained on DFDC~\cite{Dolhansky2020TheDD} dataset and evaluated on the remaining three datasets.}
  \resizebox{.5\linewidth}{!}{%
    \begin{tabular}{l|c|c|c|c|c|c|c}
    \toprule
    \rowcolor[rgb]{ 0,  0,  0} \multicolumn{7}{c|}{\textcolor[rgb]{ 1,  1,  1}{\textbf{Training Dataset: DFDC}}} & \multicolumn{1}{c}{\multirow{2}[6]{*}{\cellcolor[rgb]{ 1,  1,  1}\textbf{Evaluation Dataset}}} \\
\cmidrule{1-7}    \multicolumn{1}{c|}{\multirow{2}[4]{*}{\textbf{Model}}} & \multicolumn{3}{c|}{\textbf{With Augs}} & \multicolumn{3}{c|}{\textbf{No Augs}} &  \\
\cmidrule{2-7}          & \textbf{LogLoss} & \textbf{AUC} & \textbf{ACC} & \textbf{LogLoss} & \textbf{AUC} & \textbf{ACC} &  \\
    \midrule
    Xception & 1.4046 & 58.38\% & 55.25\% & 1.8346 & 60.31\% & 53.63\% & \multirow{6}[15]{*}{\textbf{FakeAVCeleb}} \\
\cmidrule{1-7}    Res2Net-101 & 2.0891 & 59.23\% & 56.33\% & 1.6953 & 59.77\% & 55.43\% &  \\
\cmidrule{1-7}    EfficientNet-B7 & \cellcolor[rgb]{ 1,  .753,  0}\textbf{1.0800} & 65.31\% & 61.63\% & \cellcolor[rgb]{ 1,  .753,  0}\textbf{1.0920} & \cellcolor[rgb]{ 1,  .753,  0}\textbf{71.87\%} & \cellcolor[rgb]{ 1,  .753,  0}\textbf{65.40\%} &  \\
\cmidrule{1-7}    ViT   & 1.2515 & 59.31\% & 56.00\% & 1.1361 & 60.12\% & 57.43\% &  \\
\cmidrule{1-7}    Swin  & 1.2053 & \cellcolor[rgb]{ 1,  .753,  0}\textbf{67.81\%} & \cellcolor[rgb]{ 1,  .753,  0}\textbf{62.90\%} & 1.2668 & 63.48\% & 60.25\% &  \\
\cmidrule{1-7}    MViT  & 1.2121 & 63.46\% & 60.05\% & 1.2139 & 65.75\% & 61.30\% &  \\
\cmidrule{1-7}    ResNet-3D & 1.1114 & 63.19\% & 54.50\% & 1.2748 & 62.02\% & 56.50\% &  \\
\cmidrule{1-7}    TimeSformer & 1.1582 & 65.34\% & 62.00\% & 1.6968 & 67.80\% & 61.00\% &  \\
\midrule
    \rowcolor[rgb]{ .851,  .851,  .851} \multicolumn{1}{c}{} & \multicolumn{1}{c}{} & \multicolumn{1}{c}{} & \multicolumn{1}{c}{} & \multicolumn{1}{c}{} & \multicolumn{1}{c}{} & \multicolumn{1}{c}{} &  \\
    \midrule
    Xception & 1.1784 & 67.90\% & 61.25\% & 1.7465 & 64.95\% & 58.20\% & \multirow{6}[16]{*}{\textbf{CelebDF-V2}} \\
\cmidrule{1-7}    Res2Net-101 & 1.2293 & 83.01\% & 74.95\% & 1.1859 & 83.57\% & 72.35\% &  \\
\cmidrule{1-7}    EfficientNet-B7 & 0.8278 & 79.82\% & 70.15\% & 1.2972 & 74.27\% & 68.45\% &  \\
\cmidrule{1-7}    ViT   & \cellcolor[rgb]{ 1,  .753,  0}\textbf{0.7301} & 85.62\% & \cellcolor[rgb]{ 1,  .753,  0}\textbf{76.45\%} & 0.9351 & 73.81\% & 67.25\% &  \\
\cmidrule{1-7}    Swin  & 0.8411 & 84.36\% & 76.60\% & 1.1246 & 80.34\% & 73.20\% &  \\
\cmidrule{1-7}    MViT  & 0.8548 & \cellcolor[rgb]{ 1,  .753,  0}\textbf{87.83\%} & 71.55\% & \cellcolor[rgb]{ 1,  .753,  0}\textbf{0.7711} & \cellcolor[rgb]{ 1,  .753,  0}\textbf{84.75\%} & \cellcolor[rgb]{ 1,  .753,  0}\textbf{76.75\%} &  \\
\cmidrule{1-7}    ResNet-3D & 0.7638 & 79.60\% & 72.00\% & 0.7806 & 77.88\% & 72.00\% &  \\
\cmidrule{1-7}    TimeSformer & 1.0558 & 76.48\% & 71.00\% & 1.3635 & 79.60\% & 74.00\% &  \\
    \midrule
    \rowcolor[rgb]{ .851,  .851,  .851} \multicolumn{1}{c}{} & \multicolumn{1}{c}{} & \multicolumn{1}{c}{} & \multicolumn{1}{c}{} & \multicolumn{1}{c}{} & \multicolumn{1}{c}{} & \multicolumn{1}{c}{} &  \\
    \midrule
    Xception & 1.3048 & 64.59\% & 60.95\% & 1.4530 & 67.29\% & 62.70\% & \multirow{6}[16]{*}{\textbf{FaceForensics++}} \\
\cmidrule{1-7}    Res2Net-101 & 2.7490 & 61.15\% & 56.48\% & 2.3375 & 65.15\% & 60.78\% &  \\
\cmidrule{1-7}    EfficientNet-B7 & \cellcolor[rgb]{ 1,  .753,  0}\textbf{1.2085} & 68.82\% & 64.08\% & 1.4401 & \cellcolor[rgb]{ 1,  .753,  0}\textbf{70.71\%} & \cellcolor[rgb]{ 1,  .753,  0}\textbf{65.83\%} &  \\
\cmidrule{1-7}    ViT   & 1.6779 & 66.43\% & 59.13\% & 1.6781 & 60.19\% & 57.28\% &  \\
\cmidrule{1-7}    Swin  & 1.6806 & 68.32\% & 63.65\% & 1.7493 & 64.66\% & 60.20\% &  \\
\cmidrule{1-7}    MViT  & 1.6317 & 65.82\% & 61.55\% & 1.6911 & 67.54\% & 62.88\% &  \\
\cmidrule{1-7}    ResNet-3D & 1.5308 & 57.27\% & 57.86\% & \cellcolor[rgb]{ 1,  .753,  0}\textbf{1.3507} & 58.91\% & 57.50\% &  \\
\cmidrule{1-7}    TimeSformer & 1.2122 & \cellcolor[rgb]{ 1,  .753,  0}\textbf{70.17\%} & \cellcolor[rgb]{ 1,  .753,  0}\textbf{66.79\%} & 1.9148 & 67.90\% & 60.00\% &  \\
    \bottomrule
    \bottomrule
    \end{tabular}%
    }
  \label{tab:inter-dfdc}%
\end{table}%

\clearpage
\section{Analysis of results achieved by self-supervised models}

\begin{figure*}[h!]
  \centering
  % \vspace*{-0.7cm}
  \includegraphics[width=.85\linewidth]{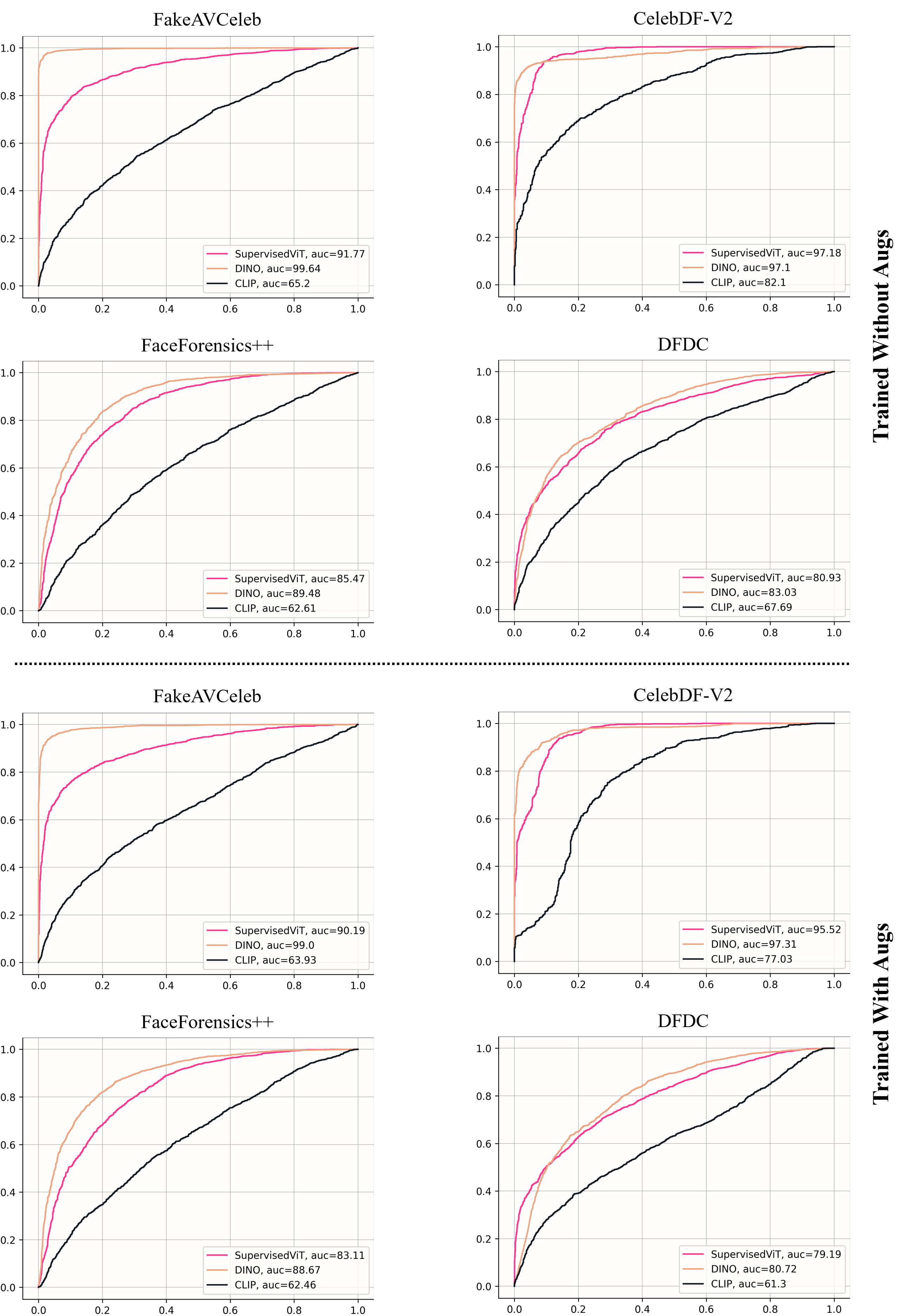}
\caption{ROC curves of self-supervised models trained and evaluated on each dataset using the intra-dataset evaluation scheme.}
\label{fig:selfsupervised_roc}
\end{figure*}

% \clearpage
% \section{Feature Visualisations}
% \begin{figure*}[!htb]
%   \centering
%   \includegraphics[width=\linewidth]{figs/cbams.png}
% \caption{Grad-CAM (Class Activation Map) visualisations of the supervised image models.}
% \label{fig:cbams}
% \end{figure*}

%%%%%%%%%%%%%%%%%%%%%%%%%%%%%%%%%%%%%%%%%%
\clearpage
%%%%%%%%%%%%%%%%%%%%%%%%%%%%%%%%%%%%%%%%%%


% \appendix









\end{document}